\documentclass{article}

\PassOptionsToPackage{numbers, compress}{natbib}

\usepackage[preprint]{neurips_2026}

\usepackage{threeparttable}
\usepackage[utf8]{inputenc} %
\usepackage[T1]{fontenc}    %
\usepackage{hyperref}       %
\usepackage{url}            %
\usepackage{booktabs}       %
\usepackage{amsfonts}       %
\usepackage{nicefrac}       %
\usepackage{microtype}      %
\usepackage{xcolor}         %
\usepackage{times}
\usepackage{epsfig}
\usepackage{graphicx}
\usepackage{amsmath,amssymb,amsfonts}
\usepackage{booktabs}
\usepackage{multirow}
\usepackage{makecell}
\usepackage{array}
\usepackage{tabularx}
\usepackage{threeparttable}
\usepackage{xspace}
\usepackage{enumitem}
\usepackage{caption}
\usepackage{subcaption}
\usepackage{siunitx}
\usepackage{colortbl}
\usepackage{pgf}
\usepackage{tikz}
\usepackage{makecell}
\usepackage{etoc}
\usepackage{float}
\usepackage{ragged2e}   %

\definecolor{lowcolor}{RGB}{245,200,200}   %
\definecolor{midcolor}{RGB}{255,245,200}   %
\definecolor{highcolor}{RGB}{180,220,180}  %
\usepackage{ifthen}

\definecolor{deltared}{RGB}{220,80,80}
\definecolor{deltagreen}{RGB}{80,170,100}
\definecolor{deltawhite}{RGB}{255,255,255}

\newcommand{\deltacell}[1]{%
  \pgfmathtruncatemacro{\deltapct}{min(100,abs(#1)*130)}%
  \ifdim #1pt < 0pt
    \edef\tempcolor{deltared!\deltapct!deltawhite}%
  \else
    \edef\tempcolor{deltagreen!\deltapct!deltawhite}%
  \fi
  \expandafter\cellcolor\expandafter{\tempcolor}%
  #1%
}

\newcommand{\heatcell}[2][]{%
  \pgfmathsetmacro{\clamped}{min(1, max(0, #2))}%
  \pgfmathtruncatemacro{\blendpct}{\clamped*100}%
  \ifnum\blendpct>70
    \pgfmathtruncatemacro{\upperpct}{min(100, (\blendpct-70)*2)}%
    \edef\tempcolor{highcolor!\upperpct!midcolor}%
  \else
    \pgfmathtruncatemacro{\lowerpct}{min(100, \blendpct*2)}%
    \edef\tempcolor{midcolor!\lowerpct!lowcolor}%
  \fi
  \expandafter\cellcolor\expandafter{\tempcolor}%
  \ifx\\#1\\#2\else\makecell{#2\\{\scriptsize$\pm$#1}}\fi%
}
\newcommand{\noheatcell}[2][]{%
  \ifx\\#1\\#2\else\makecell{#2\\{\scriptsize$\pm$#1}}\fi%
}

\newcommand{\heatcellcd}[2][]{%
  \pgfmathtruncatemacro{\blendpct}{100 - (#2/0.06)*100}%
  \ifnum\blendpct<0 \def\blendpct{0}\fi
  \ifnum\blendpct>100 \def\blendpct{100}\fi
  \ifnum\blendpct>31
    \pgfmathtruncatemacro{\upperpct}{(\blendpct-30)*2}%
    \edef\tempcolor{highcolor!\upperpct!midcolor}%
  \else
    \pgfmathtruncatemacro{\lowerpct}{\blendpct*2}%
    \edef\tempcolor{midcolor!\lowerpct!lowcolor}%
  \fi
  \expandafter\cellcolor\expandafter{\tempcolor}%
  \ifx\\#1\\%
    #2%
  \else
    \begin{tabular}{@{}c@{}}#2\\[-0.5ex]\scriptsize $\pm$#1\end{tabular}%
  \fi
}

\usepackage{pifont}
\usepackage{hyperref}
\usepackage{url}
\usepackage{balance}
\usepackage{wrapfig}
\usepackage{microtype}
\usepackage{subcaption}
\usepackage{float}  %
\usepackage{etoolbox}
\hypersetup{
    colorlinks=true,
    linkcolor=blue,
    citecolor=blue,
    urlcolor=blue
}

\newcommand{\gpt}{GPT 5.4 }
\newcommand{\gemini}{Gemini 3.1 Pro }
\newcommand{\claude}{Claude Opus 4.7 }
\newcommand{\benchmark}{\textsc{CadBench}\xspace}
\newcommand{\cadfit}{CADFit\xspace}
\newcommand{\cadrecode}{CAD-Recode\xspace}
\newcommand{\cadevolve}{CADEvolve\xspace}
\newcommand{\cadrille}{Cadrille\xspace}
\newcommand{\cadcoder}{CAD-Coder\xspace}

\newcommand{\cmark}{\ding{51}}   %
\newcommand{\xmark}{\ding{55}}   %

\title{\benchmark: A Unified Benchmark for Multimodal CAD Program Generation}

\title{\benchmark: A Multimodal Benchmark for AI-Assisted CAD Program Generation}

\author{%
  Anna C. Doris \quad
  Jacob Thomas Sony \quad
  Ghadi Nehme \quad
  Era Syla \quad
  Amin Heyrani Nobari \\
  \textbf{Faez Ahmed} \\[0.5 em]
  Massachusetts Institute of Technology \\
}

\begin{document}
\maketitle

\begin{abstract}
Recovering editable CAD programs from images or 3D observations is central to AI-assisted design, but progress is difficult to measure because existing evaluations are fragmented across datasets, modalities, and metrics. We introduce CADBench, a unified benchmark for multimodal CAD program generation. CADBench contains 18,000 evaluation samples spanning six benchmark families derived from DeepCAD, Fusion 360, ABC, MCB, and Objaverse; five input modalities including clean meshes, noisy meshes, single-view renders, photorealistic renders, and multi-view renders; and six metrics covering geometric fidelity, executability, and program compactness. STEP-based families are stratified by B-rep face count and all families are diversity-sampled to support controlled analysis across complexity and object variation. We benchmark eleven CAD-specialized and general-purpose vision-language systems, generating more than 1.4 million CAD programs. Under idealized inputs, specialized mesh-to-CAD models substantially outperform code-generating VLMs, which remain far from reliable CAD program reconstruction. CADBench further reveals three recurring failure modes: reconstruction quality degrades with geometric complexity, CAD-specialized models can be brittle under modality shift, and model rankings change across metrics. Together, these results position CADBench as a diagnostic testbed for measuring progress in editable 3D reconstruction and multimodal CAD understanding. The benchmark is publicly available at \url{https://github.com/anniedoris/CADBench}.

\end{abstract}

\section{Introduction}

Computer-aided design (CAD) models encode solid geometry in structured, parametric programs, enabling precise, editable control over design form in engineering workflows. However, creating such models remains time-consuming and expertise-intensive: prior studies estimate that even experienced engineers may require on the order of weeks to construct detailed CAD models for real-world components such as aerospace parts~\cite{roy2001quantitative}. This cost has motivated significant interest in AI systems that generate editable CAD programs directly from user inputs~\cite{zhang2026large}. Unlike approaches that reconstruct only final 3D geometry, such as meshes, voxels, or point clouds, CAD program generation aims to produce outputs that can be modified, re-parameterized, and integrated into design and manufacturing pipelines. Recent work has explored CAD generation from diverse inputs, including text descriptions~\cite{du2024blenderllm, khan2024text2cad, xie2025text}, 2D images or drawings~\cite{elistratov2026cadevolve, doris2026cad, yu2025gencad, li2025caddreamer}, and 3D observations such as meshes or point clouds~\cite{kolodiazhnyi2025cadrille, rukhovich2025cad, alam2024gencad, xu2024cad, karadeniz2025micadangelo, khan2024cad}. These approaches span general-purpose multimodal models~\cite{makatura2023can, picard2025concept}, fine-tuned language models~\cite{du2024blenderllm, doris2026cad}, and CAD-specific architectures~\cite{alam2024gencad, para2021sketchgen}.

Despite rapid progress in AI-driven CAD generation, evaluation has not kept pace. Existing CAD generation methods are typically assessed on narrow datasets or test splits, such as DeepCAD~\cite{wu2021deepcad}, Fusion 360 Gallery~\cite{willis2021fusion}, and MCB~\cite{kim2020large}. Methods are often assessed on different input modalities and metrics, making direct comparison between approaches difficult. Current benchmarks also rarely control for complexity and diversity, often overestimating the performance of CAD systems on relatively simple or uniform examples. They frequently rely on clean, CAD-derived inputs, such as idealized renders or exact geometry, failing to quantify how well models perform in scenarios more reflective of downstream CAD reconstruction tasks. Finally, existing evaluations are often limited to one or a few notions of accuracy, even though CAD program quality depends on multiple aspects of geometric fidelity, program executability, and program compactness. As a result, existing evaluations provide limited diagnostic insight into three central questions: how performance changes with increasing geometric complexity, how robust models are to modality shifts, and how model rankings and failure modes vary across CAD-specific evaluation metrics.

\begin{figure}[t]
    \centering
    \includegraphics[width=1.0\linewidth]{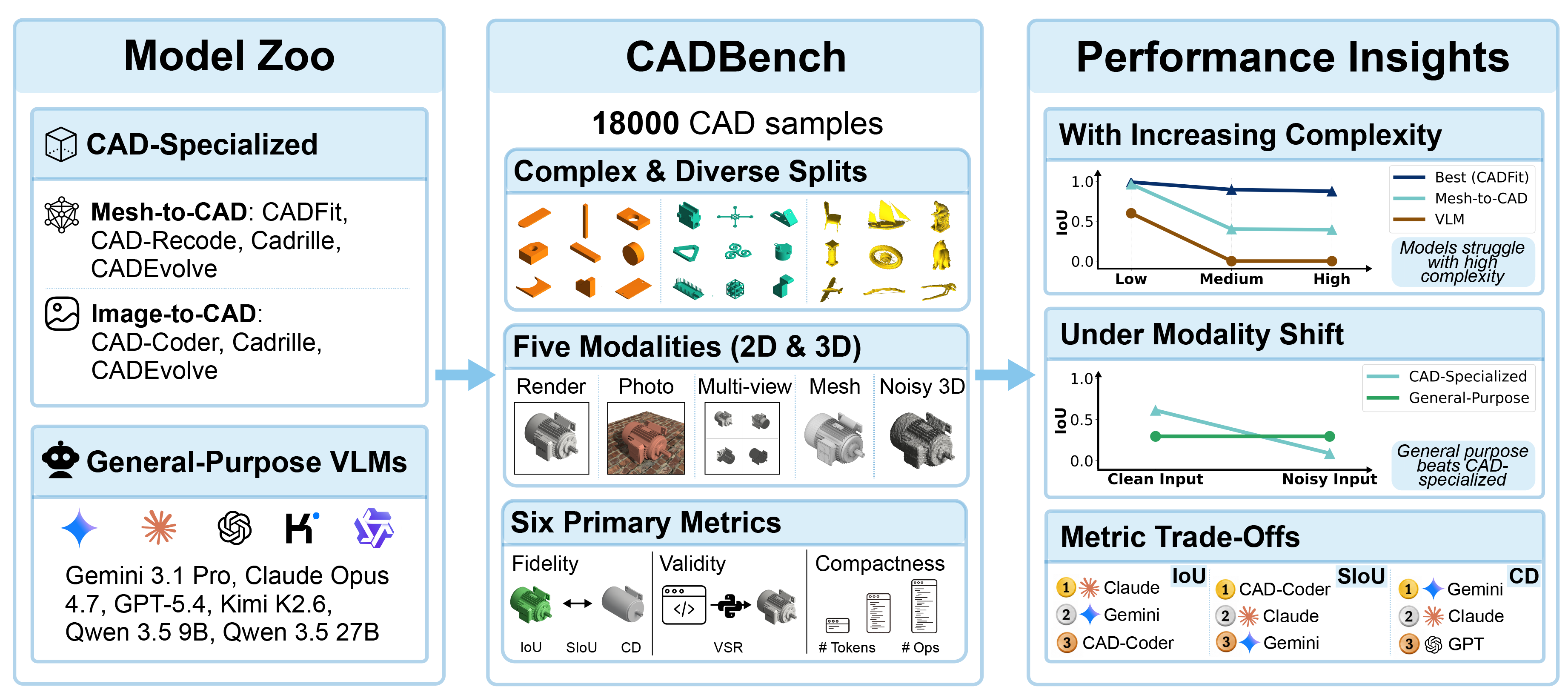}
    \caption{\textbf{Overview of \benchmark.} \benchmark evaluates eleven CAD-specialized and general-purpose models for CAD program generation across 18{,}000 CAD samples, six benchmark families, five input modalities, and six primary metrics. The benchmark is designed to support diagnostic analysis along three axes: performance with increasing geometric complexity, robustness to modality shift, and metric-dependent trade-offs.}
    \label{fig:overview_figure}
\end{figure}

We introduce \benchmark (Figure~\ref{fig:overview_figure}), a unified benchmark for CAD program generation comprising 18{,}000 samples across six benchmark families, five 2D and 3D input modalities, and six primary evaluation metrics. \benchmark addresses the gaps in current CAD evaluation through three core components: complexity- and diversity-sampled splits for controlled analysis across geometric difficulty and benchmark families; clean and noisy input modalities for measuring robustness to downstream reconstruction settings; and a metric suite that jointly captures volumetric fidelity, surface alignment, local geometric error, executability, and program compactness. Using \benchmark, we evaluate eleven CAD-specialized and general-purpose models across more than 1.4 million generations, revealing how current systems degrade with increased geometric complexity, respond to modality shifts, and vary across evaluation metrics. Together, these design choices enable a diagnostic evaluation of CAD generation systems. Our contributions are:

\begin{itemize}[leftmargin=*,topsep=2pt,itemsep=1pt,parsep=0pt]
    \item We introduce \benchmark, a multimodal benchmark for CAD program generation comprising \textbf{18{,}000 samples} across six benchmark families, five input modalities, and six evaluation metrics.
    \item We construct complexity- and diversity-aware evaluation splits that enable controlled analysis of model performance across geometric complexity and benchmark families.
    \item We incorporate clean and noisy 2D and 3D input modalities, including rendered images, photorealistic images, multi-view images, clean meshes, and noisy meshes, enabling evaluation of model robustness to modality shift.
    \item We define a multi-dimensional metric suite spanning volumetric IoU, surface IoU, Chamfer distance, valid shape rate, token count, and operation count, capturing complementary aspects of CAD generation performance.
    \item We conduct a large-scale empirical study of eleven CAD-specialized and general-purpose models across more than \textbf{1.4 million generations}, identifying key failure modes including degradation with geometric complexity, sensitivity to modality shift, and metric-dependent differences in model performances.
\end{itemize}

\section{Related Work}
Prior work has explored related domains such as text-to-CAD generation, including BlenderLLM~\cite{du2024blenderllm}, Text2CAD~\cite{khan2024text2cad}, and CADPrompt~\cite{alrashedy2024generating}. While these methods generate CAD models from natural language descriptions, engineering design is inherently geometric and visual, and many downstream workflows begin from 2D or 3D observations. Moreover, text-to-CAD is often underspecified, whereas image-to-CAD and mesh-to-CAD reconstruction can be evaluated against objective geometric ground truth. In this work, we therefore focus on tested settings in which models reconstruct executable CAD programs from image or mesh inputs.

\paragraph{Complexity and Diversity of Existing CAD Benchmarks}
\label{sec:existing_data} In the absence of standardized benchmarks for modality-to-CAD program generation, prior work has typically evaluated models on dataset-specific test splits from existing CAD datasets, including DeepCAD~\cite{wu2021deepcad}, Fusion 360 Gallery~\cite{willis2021fusion}, MCB~\cite{kim2020large}, and CC3D~\cite{cherenkova2020pvdeconv}. These datasets have enabled substantial progress, particularly when paired CAD construction sequences are available for training. Table~\ref{tab:benchmark_comparison_summary} summarizes commonly used CAD benchmarks and reports dataset-level measures of geometric complexity and visual diversity.

DeepCAD~\cite{wu2021deepcad}, Fusion 360 Gallery~\cite{willis2021fusion}, and Omni-CAD~\cite{xu2024cad} datasets provide human-generated CAD samples, but CAD operations are restricted to sketch-and-extrude only, limiting both the complexity of samples and types of operations represented in evaluation. MCB~\cite{kim2020large} contains realistic mechanical components, but evaluation on this dataset remains constrained to a fixed set of 68 object categories. Other more operation-rich and geometrically diverse CAD datasets, such as CC3D~\cite{cherenkova2020pvdeconv} and ABC~\cite{koch2019abc}, have often been overlooked for benchmarking, perhaps since ground truth design histories are not directly available. CADBench’s primary benchmark should be independently reproducible from released artifacts without third-party institutional data-use negotiations, which is why we exclude CC3D. Details are provided in Appendix D.

This reveals a clear gap: existing benchmarks are typically tied to a single dataset source, constraining evaluation to particular CAD operation types, complexity ranges, or object categories. This is misaligned with downstream needs, where CAD generation models must operate across diverse and complex generation problems. Moreover, existing benchmarks do not curate splits explicitly for complexity and diversity, limiting their diagnostic value for identifying model failure modes.

\begin{table}[t]
    \centering
    \setlength{\tabcolsep}{1.8pt}
\renewcommand{\arraystretch}{1.05}
    \small
    \footnotesize{\caption{Comparison of \benchmark against existing, commonly used benchmarks for assessing CAD generating model performance. Notably, \benchmark has higher complexity (as measured by average face count) and diversity (as measured by pairwise cosine similarity); see Appendix \ref{app:related_work} for details on these metrics. Many existing benchmarks are test splits of existing datasets, so number of samples corresponds to test set size. For existing datasets, we report the 2D modalities, 3D modalities, and reconstruction metrics provided in the original dataset; works that extend these datasets to additional modalities are discussed in Section~\ref{sec:related_modalities}.
    \label{tab:benchmark_comparison_summary}}}
    
    \resizebox{\textwidth}{!}{%
    \begin{tabular}{l c c c c c l l l}
        \toprule
        \multirow{2}{*}{\textbf{Benchmark}} & \multirow{2}{*}{\textbf{Samples}} & \multirow{2}{*}{\makecell{\textbf{Complexity}\\\textbf{Splits}}} & \multirow{2}{*}{\makecell{\textbf{Avg. Face}\\\textbf{Complexity} $\uparrow$}} & \multirow{2}{*}{\makecell{\textbf{Diversity}\\\textbf{Sampling}}} & \multirow{2}{*}{\makecell{\textbf{Avg.}\\\textbf{Similarity $\downarrow$}}} & \multirow{2}{*}{\makecell{\textbf{2D}\\\textbf{Modalities}}} & \multirow{2}{*}{\makecell{\textbf{3D}\\\textbf{Modalities}}} & \multirow{2}{*}{\makecell{\textbf{Reconstruction}\\\textbf{Metrics}}} \\
        & & & & & & & & \\
        \midrule
        \textbf{\makecell[l]{CADBench\\(Ours)}} & 18{,}000 & \cmark & 118* & \cmark & 0.428  & \makecell[l]{SV render\\MV render\\Photo. Render} & \makecell[l]{STEP$^{**}$\\Mesh\\Noisy Mesh} & \makecell[l]{IoU, SIoU\\CD, VSR\\Token Count, Op. Count} \\
        \midrule
        DeepCAD~\cite{wu2021deepcad}          & 8{,}052   & \xmark & 13 & \xmark & 0.533 & None & \makecell[l]{STEP$^{**}$\\Mesh} & \makecell[l]{$\text{ACC}_{cmd}$, $\text{ACC}_{param}$,\\CD, IR} \\[1pt]
        Fusion 360~\cite{willis2021fusion}        & 1{,}725   & \xmark & 16 & \xmark & 0.456 & None & PC      & \makecell[l]{IoU, Exact Reconst.,\\Concise.} \\[1pt]
        MCB~\cite{kim2020large}        & 11{,}700   & \xmark & N/A* & \xmark & 0.503 & None & Mesh & None \\[1pt]

        CC3D~\cite{cherenkova2020pvdeconv}        & 5000   & \xmark & N/A$^{\dagger}$ & \xmark & N/A$^{\dagger}$ & None & \makecell[l]{STEP\\3D Scan} & Chamfer Distance  (CD) \\[1pt]

        Omni-CAD~\cite{xu2024cad}          & $\sim$27K   & \xmark & 26 & \xmark & 0.483 & Multi-view Images & \makecell[l]{STEP$^{**}$ \\Point Cloud} & \makecell[l]{CD, F-score, Normal Consistency, \\ Segment Error (SegE), \\ Dangling Edge Length (DangEL), \\ Self-Intersection Ratio (SIR), \\ Flux Enclosure Error (FluxEE)} \\[1pt]
        
        \bottomrule
    \end{tabular}%
    } %
        {\footnotesize
    \noindent\parbox{\textwidth}{%
    \textit{Notes.}
    $^{*}$ Face counts are reported only for STEP-based splits and datasets, since face count is not a meaningful measure of complexity for mesh-based data.
    $^{**}$ STEP files are only available in CADBench for splits that originally provide STEP geometry.
    $^{\dagger}$ CC3D is listed as related benchmark context but is not included in CADBench aggregate metrics because access requires a separate institutional license agreement. To keep CADBench directly rerunnable from released artifacts, we restrict the primary benchmark suite to sources for which we can provide reviewer-accessible artifacts or deterministic acquisition scripts consistent with the original licenses.%
    }}
    
\end{table}

\paragraph{Modalities in Existing CAD Benchmarks}
\label{sec:related_modalities} To measure image-to-CAD reconstruction, several works have augmented the DeepCAD, Fusion 360, and MCB datasets with image modalities. For example, \cite{alam2024gencad, wang2025cad} use single-view grayscale renders of CAD solids, while \cite{kolodiazhnyi2025cadrille, elistratov2026cadevolve} generate multi-view renders tailored for specific model training paradigms. Evaluation on realistic imagery is less common. For instance, \cite{doris2026cad} evaluates generalizability using a small set of five real photographs, while \cite{li2025caddreamer} trains and evaluates on synthetic RGB renderings with textures and backgrounds, and additionally tests on real photographs of physical objects to assess generalization to real-world imagery.

For 3D-to-CAD reconstruction, a number of studies, including \cite{rukhovich2025cad, karadeniz2025micadangelo, khan2024cad, uy2022point2cyl, dupont2024transcad}, use point clouds derived from ground-truth CAD solids from datasets such as DeepCAD and Fusion 360. However, evaluation on real or noisy data remains uncommon. \cite{cherenkova2020pvdeconv} provides a dataset of over 50,000 models virtually scanned to simulate realistic sensor noise, and \cite{yu2025gencad} uses real scans of 3D-printed parts to investigate performance on real-world inputs. Furthermore, Omni-CAD~\cite{xu2024cad} introduces a large-scale multimodal dataset of over 400,000 samples, integrating text, multi-view images, and point clouds, while also evaluating model adaptability to degraded inputs such as noisy or cropped point clouds.

Overall, current 2D- and 3D-to-CAD evaluations rely heavily on idealized inputs, typically CAD-derived renders, point clouds, or meshes. Comprehensive evaluation on modalities characteristic of downstream real-world deployment, such as photographs or 3D scans, remains limited.

\paragraph{Metrics for CAD Benchmarking} In most existing works on generative models for CAD, as well as the benchmarks used to test them, performance metrics focus mostly on geometric similarity-- namely, Chamfer Distance~(CD) and Intersection over Union~(IoU)~\cite{alam2024gencad,wu2021deepcad,kolodiazhnyi2025cadrille,elistratov2026cadevolve,rukhovich2025cad,doris2026cad}. Moreover, these geometric comparison metrics may not fully encompass how well a target geometry is being reconstructed, as such macro-scale metrics simply lack the fidelity to capture fine details~(e.g., fillets on edges). Most notably, the approach used to measure these metrics is inconsistent across publications. Studies use varying numbers of points sampled on the geometry to measure CD, or employ different alignment strategies before measurement. For example, some align bounding box corners~\cite{yu2025gencad}, some align the centroids of bounding boxes~\cite{kolodiazhnyi2025cadrille,elistratov2026cadevolve,rukhovich2025cad}, and more recently, others use continuous Procrustes analysis to align~\cite{doris2026cad}. Together, these limitations motivate a unified evaluation protocol with complementary CAD-specific metrics and consistent measurement procedures.

\section{Methods}
\label{sec:benchmark_construction}

\subsection{Benchmark Construction}
Benchmark construction proceeds by selecting source datasets, stratifying and sampling shapes to control complexity and diversity, and generating standardized input modalities.

\paragraph{Benchmark Families} \benchmark is constructed to evaluate CAD program generation across both CAD reconstruction settings and more challenging out-of-distribution geometry. We therefore curate six benchmark families from existing CAD and 3D object datasets, spanning sketch-and-extrude CAD programs, richer CAD operation histories, mechanical components, and everyday object-like geometry (Figure \ref{fig:splits}). Pre-processing steps for each subcategory are detailed in Appendix \ref{app:bench_creation}. These steps establish the initial corpus, after which a multi-stage pipeline for complexity stratification and diversity-aware sampling is applied to finalize the benchmark splits.

\begin{figure}[t]
    \centering
    \includegraphics[width=\linewidth]{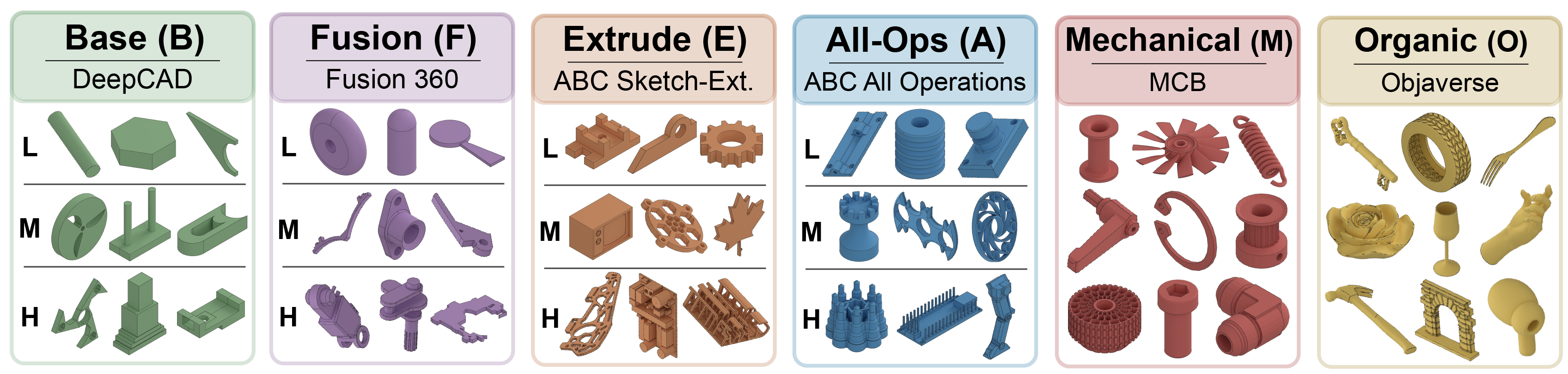}
    \caption{\textbf{Overview of \benchmark families and complexity splits.} 
    \benchmark is organized into six benchmark families: CAD-Base (B), CAD-Fusion (F), CAD-Extrude (E), CAD-All-Ops (A), CAD-Mechanical (M), and CAD-Organic (O). STEP-based families (B, F, E, A) are stratified into low- (L), medium- (M), and high-complexity (H) splits using B-rep face count, while mesh-based families (M, O) are sampled without face-count stratification because mesh face counts primarily reflect tessellation density rather than semantic CAD complexity. Complexity and diversity metrics for each split can be found in Appendix \ref{app:related_work}.}
    \label{fig:splits}
    \vspace{-3mm}
\end{figure}

\paragraph{Complexity Stratification and Diversity-Aware Sampling}  
We partition the STEP-based benchmark families (B, F, E, A) into Low (L), Medium (M), and High (H) complexity splits using B-rep face count as a proxy for geometric complexity. Face count provides a consistent measure across STEP-based datasets and reflects the number of surfaces needed to define a part; prior work identifies it as one of the geometry-based measures most strongly correlated with expert-perceived CAD modeling complexity~\cite{contero2023quantitative}. While initial splits are established by partitioning the logarithmic range of face counts into three equal intervals, we iteratively adjust the thresholds to account for right-skewed distributions, ensuring each bin contains at least 1,000 samples. This stratification is omitted for datasets with only ground-truth meshes (M, O), where face count primarily reflects tessellation density rather than underlying geometric complexity.

To promote geometric diversity, we employ a sampling pipeline using embeddings extracted via the procedure in Appendix~\ref{app:dino_embeddings}. For subcategories with complexity stratification (B, F, E, A), we perform $k$-means clustering ($k=1,000$) within each complexity tier, totaling 3,000 samples per group. For the remaining groups (M, O), we apply $k$-means ($k=3,000$) to the entire pool. After computing the cluster centroids, we select the nearest-neighbor sample to each centroid to serve as the representative sample. Ultimately, this dual approach ensures that the benchmark is both stratified by geometric complexity and more evenly distributed across the embedding space. Complexity thresholds and diversity quantification for all subcategories are provided in Appendix \ref{app:related_work} (Table \ref{tab:benchmark_comparison}).

\paragraph{Generating Mesh and Image Input Modalities}
\label{sec:modalities}
To evaluate model capabilities across both controlled and shifted input conditions, we construct five input modalities for each sample in \benchmark. For image-based evaluation, we generate: (i) \textit{single-view grayscale renders}, which provide one standardized isometric view; (ii) \textit{multi-view grayscale renders}, which combine four isometric views into a single image; and (iii) \textit{photorealistic renders}, generated via physically based rendering to introduce variation in lighting, material, perspective, and background. For 3D evaluation, we create: (iv) \textit{clean meshes} derived from the ground-truth geometry; and (v) \textit{noisy meshes}, obtained by perturbing the clean mesh geometry. Further details regarding modality generation are provided in Appendix~\ref{app:modality_gen}.

\subsection{Evaluation Metrics}
\label{sec:metrics}

We evaluate model-generated CAD programs along three axes: \emph{geometric fidelity}, \emph{executability}, and \emph{program compactness}. Geometric metrics are computed between the predicted solid $\hat{S}$ and the ground-truth shape $S$.

\paragraph{Geometric fidelity.}
We report three complementary measures of geometric reconstruction quality. \textbf{Volumetric IoU (IoU)} measures overlap between voxelized occupancies of the predicted shape $\hat{S}$ and ground-truth shape $S$, capturing global volumetric agreement. \textbf{Chamfer Distance (CD)} is computed between point samples from the predicted and target surfaces, capturing local surface error. \textbf{Surface IoU (SIoU)} measures bidirectional surface coverage under a distance threshold: a predicted surface point is counted as matched if it lies within $\tau$ of the target surface, and vice versa. We set $\tau$ to $1\%$ of the target bounding-box diagonal and average the two coverage terms. Together, these metrics distinguish volumetric overlap, local geometric deviation, and thresholded surface alignment. We report these metrics after applying the alignment procedure described in Appendix~\ref{app:metric_details}; unaligned metrics are also available in our codebase. For benchmark splits, we report median IoU and median SIoU with failed executions counted as zero, and median CD over successfully executed programs.

\paragraph{Executability.}
We measure whether a generated CAD program can be executed to produce a valid solid. We report \textbf{Valid Shape Rate (VSR)}, the fraction of generated programs that execute successfully and yield a valid shape. Invalid predictions include programs with syntax errors, failed CAD operations, or outputs that cannot be converted into valid solid geometry for evaluation.

\paragraph{Program compactness.}
To characterize the degree to which a model reconstructs geometry using concise syntax, we report two measures: \textbf{Token Count}, the number of generated tokens, and \textbf{Operation Count}, the number of CAD operations in the generated program. 

Aggregate benchmark scores are computed as weighted averages of split-level metrics, with weights proportional to the number of samples in each split. Full mathematical definitions and implementation details for all metrics are provided in Appendix~\ref{app:metric_details}.

\subsection{Models Evaluated}

We focus on models that generate executable CAD programs from image or mesh inputs, as editable CAD outputs are indispensable for downstream engineering workflows. Consequently, we categorize our evaluation and analysis based on the input modality: i) mesh-conditioned and ii) image-conditioned models.

\paragraph{Mesh-conditioned Models}
 We define mesh-conditioned methods as models capable of receiving a mesh as input, often internally converting it into another representation, such as a point cloud or rendered views, in order to generate an editable CAD program or feature tree. For mesh-conditioned CAD program generation, we evaluate \cadfit~\cite{nehme2026cadfitprecisemeshtocadprogram}, \cadevolve~\cite{elistratov2026cadevolve}, \cadrecode~\cite{rukhovich2025cad}, and \cadrille~\cite{kolodiazhnyi2025cadrille}. We evaluate these methods because they are among the state of the art, demonstrating top-tier reconstruction performance (mean/median IoU $\geq$ 0.8) on established benchmarks such as DeepCAD and Fusion 360 Reconstruction test sets \cite{nehme2026cadfitprecisemeshtocadprogram, elistratov2026cadevolve, rukhovich2025cad, kolodiazhnyi2025cadrille}. Each mesh-conditioned model is evaluated on both clean and noisy mesh modalities using the method's default inference settings. With the exception of \cadfit, which is deterministic, each mesh-conditioned method is evaluated over three independent runs, and we report the mean and standard deviation across runs.

Although \cadrille and \cadevolve can also be run from image inputs---and we evaluate both methods on \benchmark's single-view renders---we primarily report them as mesh-conditioned methods because they achieve their strongest CADBench performance when evaluated from mesh-derived inputs. Performance degrades substantially when image inputs are not tailored to each model's native rendering format; this finding is further discussed in Section~\ref{sec:results}.

\paragraph{Image-conditioned models}
Image-conditioned models take one or more images as input and generate editable CAD feature trees. We further divide these into (1) general-purpose vision-language models (VLMs) and (2) CAD-specific image-to-CAD models.

\paragraph{General-Purpose VLMs.}
We evaluate a set of frontier proprietary VLMs, including Claude Opus 4.7, Gemini 3.1 Pro, and GPT-5.4. These models are selected as representative state-of-the-art VLMs from Anthropic, Google, and OpenAI, respectively, based on strong performance on general multimodal reasoning benchmarks (e.g., MMMU-Pro\footnote{https://artificialanalysis.ai/evaluations/mmmu-pro}). We include Kimi K2.6, an open-weight model, due to strong performance on the same benchmark. We also include open-weight models Qwen 3.5 27B and Qwen 3.5 9B, as the Qwen family has been widely adapted for multimodal and code generation tasks. For all off-the-shelf VLMs, we use recommended inference settings; full details are provided in Appendix \ref{app:methods_inference}. For each of the three image modalities, each model is evaluated over three independent runs, and we report mean and standard deviation.

\paragraph{CAD-specific models.}
For CAD-specialized, image-conditioned generation, we report \cadcoder~\cite{doris2026cad} as the primary image-conditioned CAD-specific model. We also evaluate \cadrille and \cadevolve on image inputs for completeness, but primarily categorize them as mesh-conditioned methods as explained above. Although each model was originally trained on a specific image format, we do not tailor inputs to these model-specific formats at test time. Instead, we evaluate all models using the unified set of renderings described in Section~\ref{sec:modalities}: single-view, multi-view, and photorealistic renders. \cadcoder is deterministic and run once per modality.

\section{Results and Analysis}
\label{sec:results}
\begin{table*}[t]
\centering
\setlength{\tabcolsep}{5pt}
\renewcommand{\arraystretch}{1.15}
\caption{\textbf{Aggregate CADBench performance under idealized input modalities.} Mesh-to-CAD models are evaluated on clean meshes, while image-to-CAD models are evaluated on single-view grayscale renders. Scores aggregate performance across all benchmark splits.}
\label{tab:aggregate_results}
\resizebox{\textwidth}{!}{%
\begin{tabular}{lccccccccccc}
\toprule
& \multicolumn{4}{c}{Mesh-to-CAD} & \multicolumn{7}{c}{Image-to-CAD} \\
\cmidrule(lr){2-5} \cmidrule(lr){6-12}
Metric
& \rotatebox{30}{CADFit} & \rotatebox{30}{CAD-Recode} & \rotatebox{30}{CADEvolve} & \rotatebox{30}{Cadrille}
& \rotatebox{30}{Claude Opus 4.7} & \rotatebox{30}{Gemini 3.1 Pro} & \rotatebox{30}{GPT-5.4} & \rotatebox{30}{Kimi K2.6} & \rotatebox{30}{Qwen 3.5 27B} & \rotatebox{30}{Qwen 3.5 9B} & \rotatebox{30}{CAD-Coder} \\
\midrule
IoU $\uparrow$
& \textbf{\heatcell{0.859}} & \heatcell{0.673} & \heatcell{0.707} & \heatcell{0.687} & \heatcell{0.412} & \heatcell{0.382} & \heatcell{0.158} & \heatcell{0.224} & \heatcell{0.015} & \heatcell{0.000} & \heatcell{0.354} \\
SIoU $\uparrow$
& \textbf{\heatcell{0.679}} & \heatcell{0.503} & \heatcell{0.498} & \heatcell{0.517} & \heatcell{0.108} & \heatcell{0.106} & \heatcell{0.029} & \heatcell{0.042} & \heatcell{0.002} & \heatcell{0.000} & \heatcell{0.115} \\
CD $\downarrow$
& \textbf{\heatcellcd{0.038}} & \heatcellcd{0.056} & \heatcellcd{0.062} & \heatcellcd{0.056} & \heatcellcd{0.101} & \heatcellcd{0.087} & \heatcellcd{0.109} & \heatcellcd{0.111} & \heatcellcd{0.176} & \heatcellcd{0.195} & \heatcellcd{0.157} \\
\midrule
VSR $\uparrow$
& \textbf{\heatcell{0.981}} & \heatcell{0.912} & \heatcell{0.967} & \heatcell{0.939} & \heatcell{0.798} & \heatcell{0.729} & \heatcell{0.513} & \heatcell{0.616} & \heatcell{0.335} & \heatcell{0.099} & \heatcell{0.950} \\
\midrule
Token Count
& \noheatcell{552} & \noheatcell{236} & \noheatcell{284} & \noheatcell{220} & \noheatcell{186} & \noheatcell{182} & \noheatcell{216} & \noheatcell{219} & \noheatcell{151} & \noheatcell{141} & \noheatcell{364} \\
Op Count
& \noheatcell{42} & \noheatcell{25} & \noheatcell{22} & \noheatcell{23} & \noheatcell{14} & \noheatcell{13} & \noheatcell{15} & \noheatcell{18} & \noheatcell{12} & \noheatcell{13} & \noheatcell{43} \\
\bottomrule
\end{tabular}%
}
\end{table*}

\benchmark serves both as an aggregate leaderboard and as a diagnostic benchmark for analyzing CAD generation performance. We first summarize overall model performance under clean input conditions. We then use \benchmark's comprehensive splits, modalities, and metrics to examine how performance changes with geometric complexity, how models respond to input modality shift, and whether different evaluation metrics reveal distinct failure modes.

\paragraph{How do current CAD generators perform under clean inputs?}
Table~\ref{tab:aggregate_results} summarizes overall model performance on \benchmark under clean input conditions: clean meshes for mesh-to-CAD models and single-view grayscale renders for image-to-CAD models. Detailed per-split results are provided in Appendix~\ref{app:full_results_ideal_input}. At the aggregate level, mesh-conditioned methods substantially outperform image-conditioned methods across all geometric fidelity metrics, reflecting the advantage of direct 3D observations over ambiguous single-view images. Among all models tested, \cadfit achieves the strongest reconstruction performance, with the highest IoU, the highest SIoU, the highest VSR, and the lowest CD, though the method's accuracy comes with substantially longer inference time (Appendix~\ref{app:inference_speed}).

Among image-conditioned models, performance is lower and rankings depend on the metric: \claude achieves the highest aggregate IoU, \gemini achieves the lowest Chamfer distance, and \cadcoder achieves the highest SIoU. \cadcoder also achieves the highest valid shape rate among image-conditioned methods, suggesting that CAD-specific training improves executability even if geometric fidelity remains limited. In contrast, the Qwen models achieve near-zero IoU largely because they often fail to produce valid executable shapes, indicating that general code fluency does not necessarily translate into reliable CAD syntax. Notably, the open-weight Kimi K2.6 model outperforms the closed-source \gpt model across two geometric fidelity metrics, indicating that frontier proprietary VLMs do not uniformly dominate CAD program generation.

\begin{figure}
\centering
\includegraphics[width=1\linewidth]{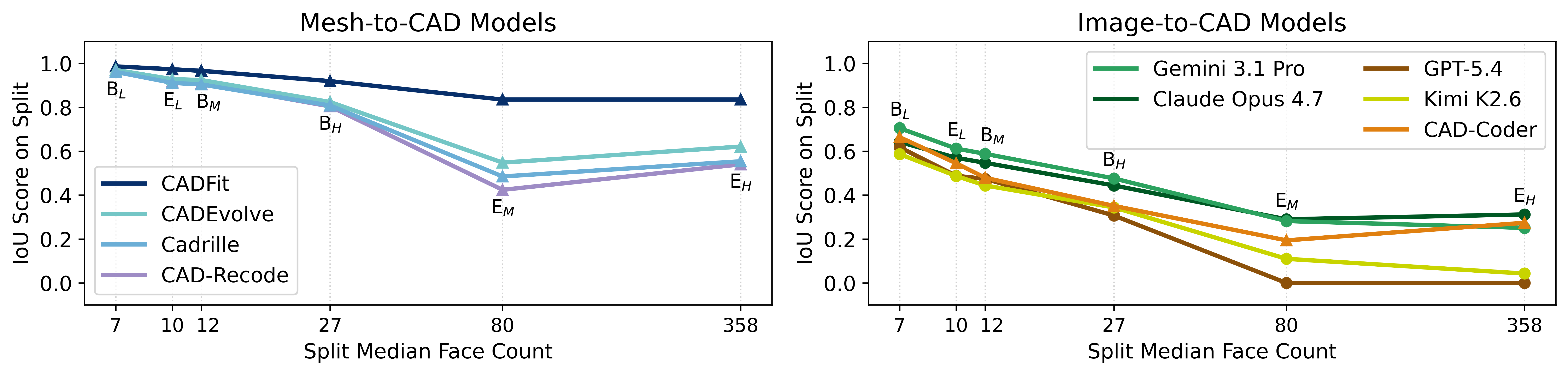}
\caption{\textbf{Model performance as a function of geometric complexity}, as measured by face count. The y-axis shows median IoU and the x-axis shows median face count for each split (on a log scale). Results are shown for Base (B) and Extrude (E) families, which include only sketch and extrude operations. Performance declines with increasing split median face count, suggesting face count captures problem difficulty.}
\label{fig:complexity_analysis}
\end{figure}

\paragraph{Does CAD reconstruction get harder as geometry becomes more complex?}
To isolate the relationship between geometric complexity and reconstruction fidelity, we analyze performance on benchmark families containing only sketch-and-extrude operations (B, E). As shown in Figure~\ref{fig:complexity_analysis}, IoU score on the split generally decreases as split median face count increases for both mesh-conditioned and image-conditioned methods. This indicates that face-count stratification captures a meaningful axis of reconstruction difficulty: models that perform well on low-face-count parts often degrade sharply on higher-face-count splits.

For the Extrude family, many models achieve similar IoU on the medium- and high-complexity splits. This suggests a possible saturation effect, where models already struggle to recover moderately complex sketch-and-extrude geometry, so additional increases in face count do not lead to proportionally lower IoU. \cadfit is the main exception, remaining comparatively robust across complexity levels. A complementary analysis including All-Ops splits is provided in Appendix~\ref{app:complexity_strat}; in this broader setting, the relationship between face count and IoU is less monotonic, suggesting that face count is useful within controlled operation families but does not fully capture reconstruction difficulty when operation vocabulary is expanded. 

A small ablation with \cadevolve further suggests that diversity-aware sampling increases benchmark difficulty: within the same face-count range, diversity-sampled splits produce lower median aligned IoU performances than randomly sampled splits, with the largest relative drop on Extrude-Medium ($-11.33\%$; Appendix~\ref{app:diversity_sampling}).

\begin{figure}[t]
    \centering
    \includegraphics[width=0.95\linewidth]{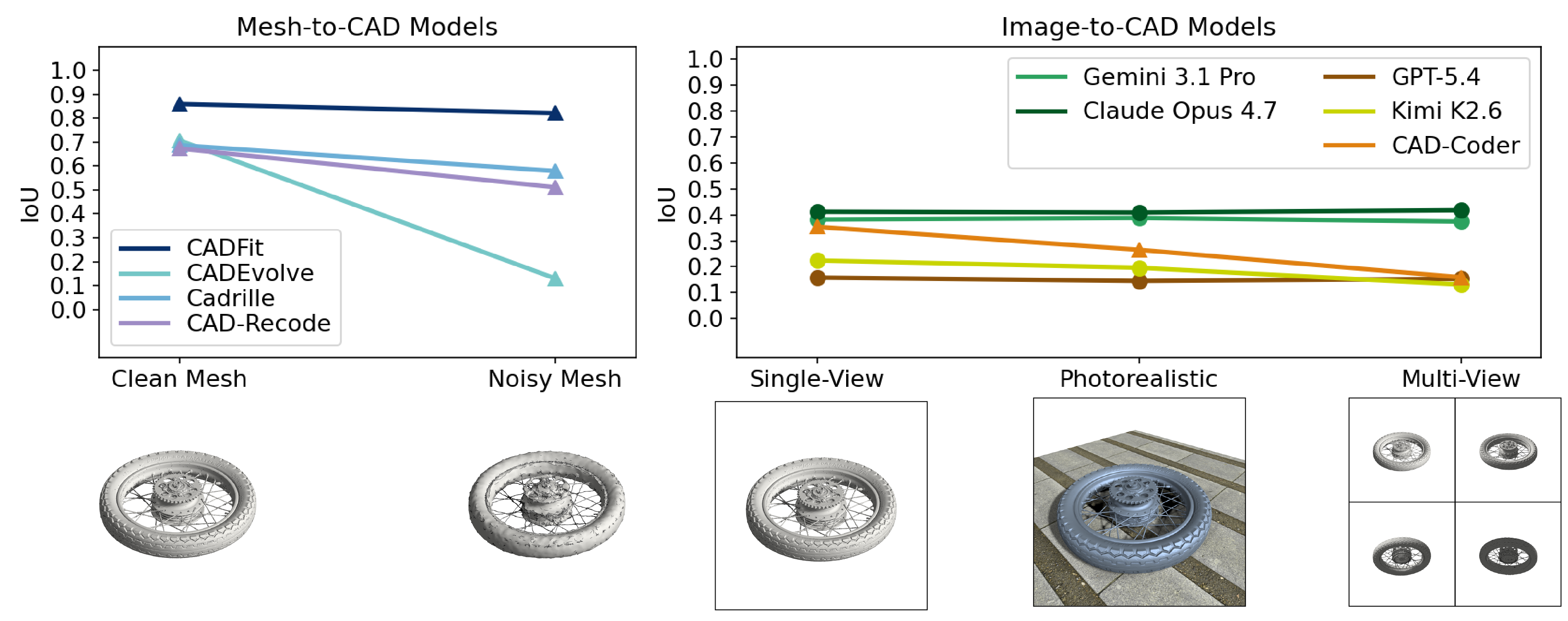}
    \caption{\textbf{Robustness to input modality shift.} Aggregate IoU is shown for mesh-to-CAD models under clean and noisy mesh inputs, and for image-to-CAD models under single-view grayscale, photorealistic, and multi-view image inputs. Examples of each input modality are also shown.}
    \label{fig:modality_comparison}
    \vspace{-3mm}
\end{figure}

\paragraph{How fragile are CAD generators to modality shift?}
Figure~\ref{fig:modality_comparison} compares model performance across clean and shifted input modalities. Mesh-conditioned CAD-specialized models are sensitive to mesh noise: CADFit, CADEvolve, Cadrille, and CAD-Recode lose 0.051, 0.577, 0.108, and 0.162 in  aggregate IoU respectively when moving from clean to noisy mesh inputs (Appendix~\ref{app:modality_shift}). The largest degradation occurs for \cadevolve, which drops from 0.707 IoU on clean meshes to 0.130 IoU on noisy meshes ($\Delta$IoU = -0.577). This may be partly explained by a compound modality-shift effect: \cadevolve first renders the input mesh into multi-view images before generating CAD code, so mesh perturbations can propagate through the rendering stage and alter the visual observations passed to the model, potentially amplifying the effect of geometric noise.

Image-conditioned CAD-specialized models also degrade under input shifts. \cadcoder drops modestly on photorealistic renders ($\Delta$IoU = -0.089), but more substantially on multi-view inputs ($\Delta$IoU = -0.195) relative to single-view grayscale renders. Because \cadcoder was trained on single-view images, this suggests that view composition is more disruptive than rendering style alone. We observe a similar sensitivity for \cadrille and \cadevolve when they are evaluated from image inputs: although both methods can accept images, performance is poor on \benchmark's standardized single-view renders, likely because these inputs differ from the image formats used during training. In contrast, general-purpose VLMs show comparatively small changes in aggregate IoU across image modalities, indicating stronger robustness to changes in lighting, material, and view layout (Appendix~\ref{app:modality_shift}).

These results reveal a trade-off in current models: CAD-specialized models achieve the strongest performance under clean or familiar inputs, but can degrade sharply under realistic modality shifts, while general-purpose VLMs are more robust to input variation but remain substantially less accurate overall. This points to the need for further work on CAD generation methods that combine geometric accuracy with robustness to downstream input conditions, characteristic of real-world applications.

\paragraph{\textbf{Are multiple geometric fidelity metrics necessary?}}
Aggregate rankings in Table~\ref{tab:aggregate_results} depend on the choice of geometric fidelity metric. Among image-conditioned models, \claude achieves the highest IoU, \gemini achieves the lowest CD, and \cadcoder achieves the highest SIoU. Although these differences are modest and all image-conditioned models remain far below the strongest mesh-conditioned methods, the change in ranking suggests that IoU, CD, and SIoU capture distinct aspects of reconstruction quality rather than providing interchangeable measurements.

To quantify this relationship, we compute sample-level Spearman rank correlations between IoU, SIoU, and CD across all syntactically valid model-generated CAD programs (Appendix~\ref{app:metrics_redundant}). The metrics exhibit the expected directional relationships but are not redundant: IoU and SIoU are only moderately correlated ($\rho=0.57$), while CD is negatively correlated with both IoU ($\rho=-0.73$) and SIoU ($\rho=-0.87$). The stronger correlation between CD and SIoU likely reflects that both are surface-based metrics, whereas IoU measures volumetric agreement. These results motivate reporting all three metrics. Appendix~\ref{app:visual_metrics} further illustrates cases where IoU and SIoU provide complementary diagnostic information: high IoU can mask failures to recover fine surface details, while high SIoU can occur when surfaces are near the target but the enclosed volume is incorrect. Such distinctions are important in CAD, where surface features such as teeth, holes, grooves, and interfaces can be critical to part function.

\section{Conclusion}
\label{sec:conclusion}

We introduced \benchmark, a unified multimodal benchmark for CAD program generation spanning six benchmark families, complexity- and diversity-sampled splits, clean and noisy input modalities, and six primary evaluation metrics. In addition to serving as an overall leaderboard, these benchmark design choices enable targeted analysis of where current CAD generators succeed and fail. The complexity-stratified splits allow us to ask how reconstruction performance changes as geometric difficulty increases, revealing that most models degrade substantially on higher-complexity shapes. The clean and noisy modality variants allow us to evaluate robustness to input shift, showing that CAD-specialized models often achieve strong performance under clean inputs but can degrade sharply under noisier or less familiar modalities, while general-purpose VLMs are more robust but remain less accurate overall. Finally, the inclusion of multiple geometric fidelity metrics allows us to test whether model rankings are metric-dependent, demonstrating that IoU, CD, and SIoU capture complementary aspects of reconstruction quality rather than interchangeable measurements. Overall, \benchmark provides a foundation for measuring progress in AI-assisted CAD reconstruction and highlights the need for future methods that combine geometric accuracy, executable and editable outputs, compact construction procedures, and robustness to realistic input conditions.

\paragraph{Limitations.}
We use ``editable CAD program'' operationally: a generated program must execute in a CAD kernel to produce a valid solid that can be inspected and manually edited as code. \benchmark does not require recovery of the original designer's feature tree, sketch constraints, construction order, parameterization, or design intent. This is unavoidable because only some source families provide human-authored construction histories, while geometry-only families provide final shapes but not the modeling processes that created them. Thus, \benchmark evaluates executable solid reconstruction, geometric fidelity, and program compactness, but cannot uniformly measure operation-level equivalence, semantic program equivalence, constraints, or correspondence to a human-authored feature tree; these remain important directions for future benchmarks. Our VLM evaluation also measures standardized direct generation rather than best-achievable performance: we use fixed prompts without extensive prompt tuning, tool use, self-correction, or multi-turn agentic refinement, all of which may improve absolute performance. Finally, our STEP-based complexity splits use B-rep face count, which is consistent but incomplete: repeated patterns such as arrays of holes can dominate high-face-count splits, making them challenging but biased toward one form of CAD complexity.

\newpage

\bibliographystyle{unsrtnat}   %
\bibliography{references}

\newpage
\appendix
\addcontentsline{toc}{part}{Appendices}

\etocsettocstyle{\large\textbf{Table of Contents for Appendices}\par\vspace{1em}}{}

\makeatletter
\begingroup
  \let\oldlsection\l@section
  \let\oldlsubsection\l@subsection
  \let\oldlsubsubsection\l@subsubsection %
  
  \renewcommand{\l@section}[2]{\oldlsection{\small #1}{\small #2}}
  \renewcommand{\l@subsection}[2]{\oldlsubsection{\small #1}{\small #2}}
  \renewcommand{\l@subsubsection}[2]{\oldlsubsubsection{\small #1}{\small #2}} %
  
  \localtableofcontents
\endgroup
\makeatother

\vspace{2em}

\newpage

\section{Additional Method Details}
\label{app:methods}
\subsection{Overview of \benchmark Families.}

\begin{table}[H]
\small
\centering
\renewcommand{\arraystretch}{1.3} %
\begin{tabular}{@{}llp{7.5cm}@{}} %
\toprule
\textbf{Subcategory} & \textbf{Source} & \textbf{Core Evaluative Focus} \\ \midrule
\textbf{CAD-Base (B)} & DeepCAD~\cite{wu2021deepcad} & \multirow{2}{7.5cm}{Establishes a standard reference using benchmarks to ensure continuity with prior research.} \\
\textbf{CAD-Fusion (F)} & Fusion 360~\cite{willis2021fusion}\cite{lambourne2021brepnet} & \\ \midrule
\textbf{CAD-Extrude (E)} & \multirow{2}{*}{ABC~\cite{koch2019abc}} & \multirow{2}{7.5cm}{Escalates difficulty via intricate sketch-extrude geometry (E) and models utilizing an expanded operation set (A)} \\
\textbf{CAD-All-Ops (A)} & & \\ \midrule
\textbf{CAD-Mechanical (M)} & MCB~\cite{kim2020large} & \multirow{2}{7.5cm}{Bridges the gap to practical application by presenting engineering components (M) and real-world objects (O).} \\
\textbf{CAD-Organic (O)} & Objaverse~\cite{deitke2023objaverse} & \\ \bottomrule
\end{tabular}
\vspace{0.5em}
\caption{Overview of \benchmark families.}
\label{tab:subcategories}
\end{table}

\subsection{Extracting Feature Vectors for Objects using DINOv3}
\label{app:dino_embeddings}

To obtain a representative feature vector for each object across our datasets, we employ a vision-based encoding pipeline inspired by the procedure utilized in \cite{ataei2026zero}. Each object is rendered into eight grayscale isometric views using \texttt{pythonOCC} and processed through a DINOv3 \cite{simeoni2025dinov3} backbone (\texttt{dinov3-vitb16-pretrain-lvd1689m}) to extract 768-dimensional latent embeddings. These vectors are then averaged to produce a single representative feature vector per object. This image-based procedure is highly effective across various 3D file formats, including STEP, STL, GLB, and OBJ.

\begin{figure}[H]
    \centering
    \includegraphics[width=\textwidth]{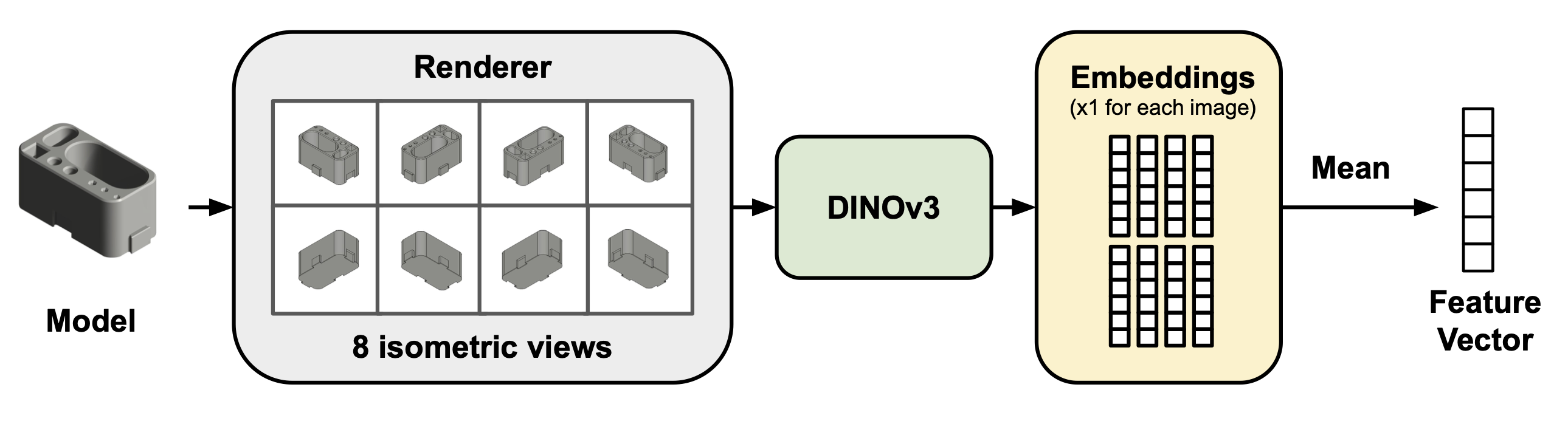}
    \caption{Procedure for extracting feature vectors for objects using DINOv3}
    \label{fig:samples_modality}
\end{figure}

\subsection{More Details about creating Bench Subcategories}
\label{app:bench_creation}

\paragraph{CAD-Base (B):} This subcategory consists of CAD Models derived from the DeepCAD dataset. We used the DeepCAD test split introduced in the GenCAD paper \cite{alam2024gencad}, consisting of 7629 samples. We first filtered the dataset to de-duplicate and to filter to only samples containing single bodies. This exclusion of multi-body designs and assemblies is motivated by the nature of the feature tree reconstruction task, which is fundamentally intended for individual parts. In standard engineering workflows, assemblies are not designed as a monolithic entity within a single feature tree. Instead, each component is designed independently with its own feature history. To align with this design intent, we filtered out models consisting of multiple bodies, leaving a total of 6210 samples. We then applied complexity-based stratification and diversity-aware sampling as described in Section \ref{sec:benchmark_construction}. 

\paragraph{CAD-Fusion (F):} This subcategory consists of CAD Models derived from the Fusion 360 Reconstruction and Segmentation datasets. Since the Fusion 360 Reconstruction test set contains only 1,725 samples, we augmented it with models from the Fusion 360 Segmentation Dataset. Following the CAD-Base filtering criteria, we excluded all multi-body designs, leaving a total of 37,184 CAD models. We then applied complexity-based stratification and diversity-aware sampling as described in Section \ref{sec:benchmark_construction}. 

\paragraph{CAD-Extrude (E) and CAD-All-Ops (A):} This subcategory consists of CAD Models derived from the ABC dataset \cite{koch2019abc}. We began by preprocessing the ABC dataset—originally comprising one million samples—using a pipeline that sought to identify and exclude objects with multiple bodies and stray surfaces. We further targeted objects containing specific geometric entities within the STEP file (such as \texttt{TRIANGULATED\_FACE}) to filter out unsuitable samples, like pseudo-STEP models, as these function merely as mesh-to-STEP wrappers and lack true geometric primitives. Utilizing the original FeatureScript ($\texttt{ofs}$) files, we further filtered out models containing text-based ($\texttt{BTMSketchTextEntity}$) or image-based ($\texttt{BTMSketchImageEntity}$) sketches. The remaining models were then divided into two distinct partitions based on their operation history: one partition ($\approx$ 126K samples) consisting of models defined by sketch ($\texttt{newSketch}$) and extrude ($\texttt{extrude}$) operations, and the other partition ($\approx$ 62K samples) comprising models that incorporate more complex features such as \texttt{fillet}, \texttt{revolve}, \texttt{chamfer}, \texttt{loft}, \texttt{sweep}, \texttt{externalThreads} and 
\texttt{internalThreads}. Finally, complexity-based stratification and diversity-aware sampling were applied to create CAD-Extrude and CAD-All-Ops, respectively.

\paragraph{CAD-Mechanical (M):} This subcategory consists of samples derived from the MCB dataset. We use the test split from the MCB Dataset, consisting of 11716 samples. We then ran a watertight check and repair, filtered to single body samples, removed duplicates, and eliminated samples for which no images were generated; this left 7106 samples. We then applied diversity-aware sampling as described in Section \ref{sec:benchmark_construction} to obtain a subset of 3000 samples. The original MCB dataset is split into 68 classes of mechanical objects. Through our diversity sampling procedure, 67/68 classes were represented in the 3000 sample benchmark. The three most represented classes are 1) conventional rivets, 2) keys and keyways, splines, and 3) screws and bolts with cylindrical heads.

\paragraph{CAD-Organic (O):} This subcategory consists of samples derived from the Objaverse dataset. While Objaverse is a large-scale collection of general 3D models, we select specific categories deemed most "CAD-suitable" based on their geometric structure and functional design:

\begin{itemize}
    \item Furniture-Home
    \item Science-Technology
    \item Architecture
    \item Cars-Vehicles
    \item Electronics-Gadgets
\end{itemize}

From an initial pool of approximately 800,000 models, filtering for these categories yielded a final subset of 226,865 samples.
To ensure geometric validity of the meshes used in our dataset, we applied a mesh repair pipeline to convert non-watertight meshes into watertight manifolds whenever possible. A mesh is considered watertight when it forms a closed 2-manifold surface with no holes or boundary edges. Watertight meshes are essential for many downstream tasks such as volume computation, physical simulation, geometry learning models, and surface reconstruction. However, many meshes in large-scale datasets such as Objaverse contain geometric defects, including holes in the surface, duplicate vertices or faces, non-manifold edges, unreferenced vertices, and degenerate triangles. To address these issues, we implemented a repair pipeline using PyMeshLab, which provides robust mesh-processing filters. We subsequently de-duplicated and filtered for single body samples, which left 20148 samples. We then applied the mesh-based dataset procedure described above to select a benchmark subset of 3000 samples.

\subsection{Generating Mesh and Image Inputs for Model Evaluation}
\label{app:modality_gen}

We describe in greater detail the methodology used to generate the mesh and image inputs for benchmarking and evaluating various models.

\begin{figure}[htbp]
     \centering
     \begin{subfigure}[b]{0.49\textwidth}
         \centering
         \includegraphics[width=\linewidth]{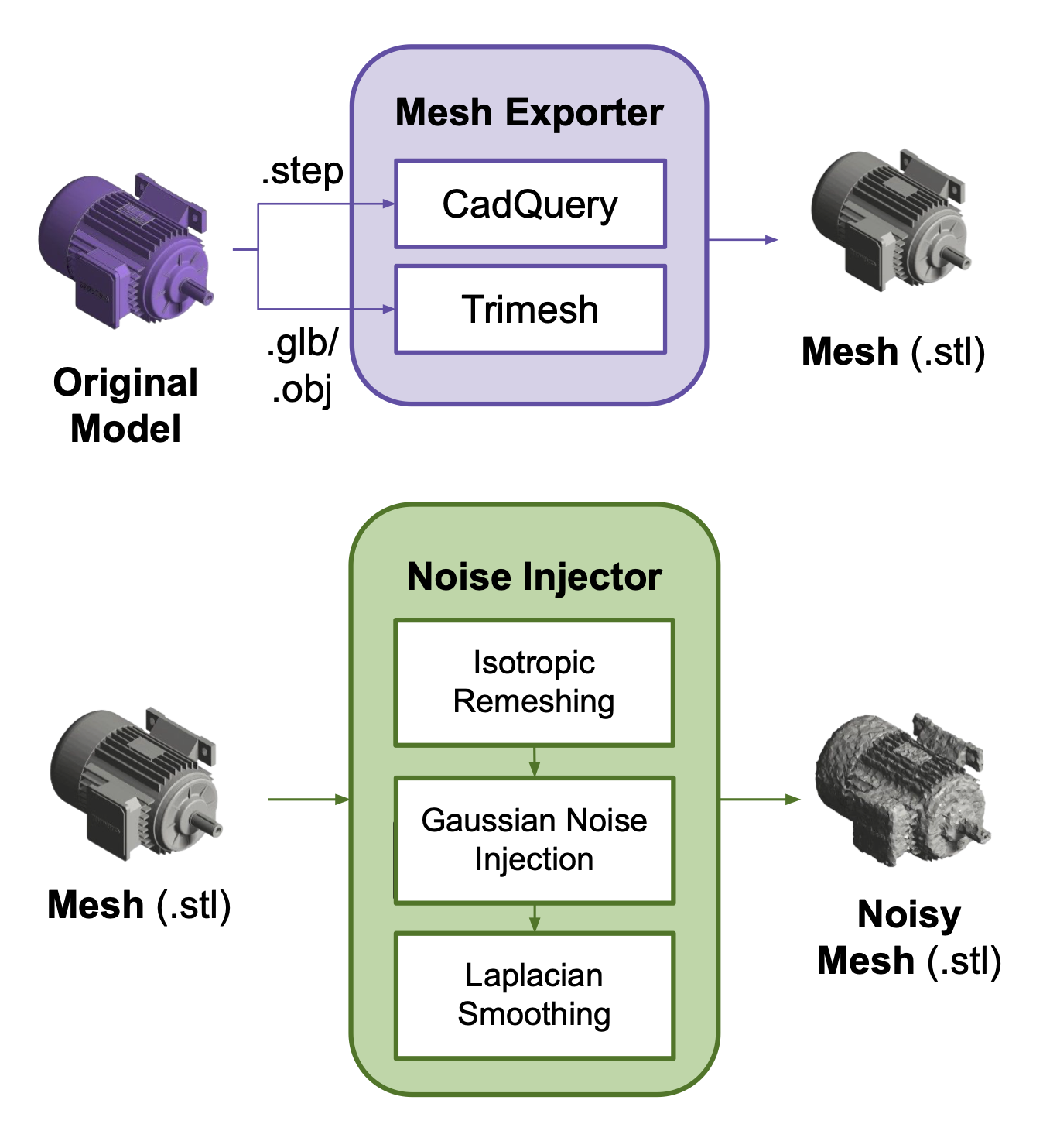}
         \caption{Methodology for generating mesh inputs}
         \label{fig:left}
     \end{subfigure}
     \hfill %
     \begin{subfigure}[b]{0.49\textwidth}
         \centering
         \includegraphics[width=\linewidth]{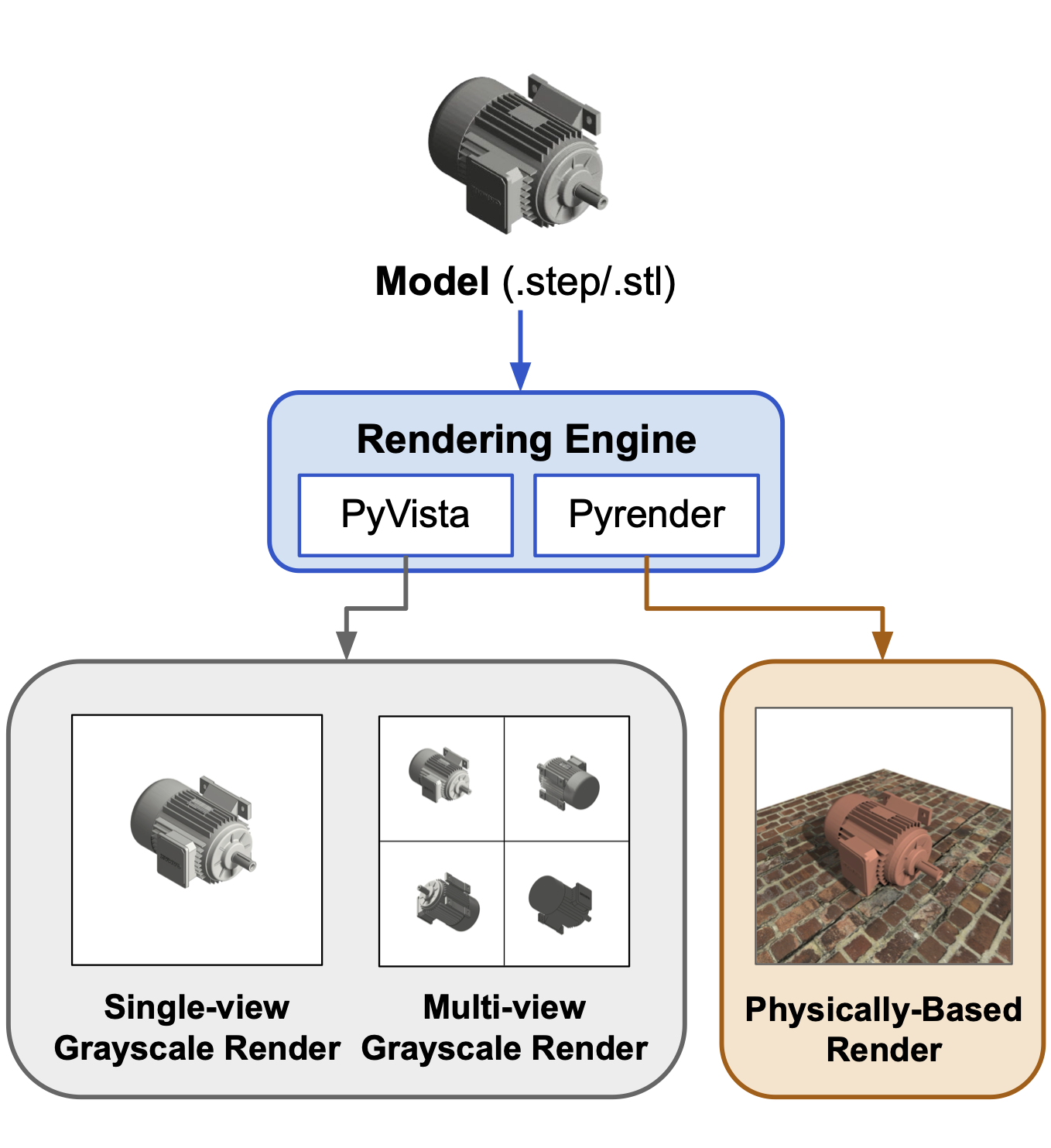}
         \caption{Methodology for generating image inputs}
         \label{fig:right}
     \end{subfigure}
     
     \caption{Schematic representation of the workflow involved in generating mesh and image inputs}
     \label{fig:together}
\end{figure}

\paragraph{Mesh Generation:} For the B, F, E, and A subcategories, we utilize \texttt{CadQuery} to export STEP models to STL format using default tessellation settings. For M and O, we import source OBJ or GLB files and perform STL conversion via the \texttt{trimesh} library. In all cases, meshes are centered at the origin and normalized to fit within a $[-1, 1]^3$ bounding box. To generate noisy variants, models undergo isotropic remeshing via \texttt{PyMeshLab}, followed by the injection of Gaussian noise ($\sigma = 0.005$) into the vertices. We subsequently apply Laplacian smoothing to suppress sharp artifacts prior to final export.

\paragraph{Grayscale Rendering:} High-resolution ($1200 \times 1200$) single-view grayscale images are generated by rendering STL models in an isometric orientation using \texttt{PyVista}. To emphasize structural definition, we apply a neutral matte finish (\texttt{\#BFBEBA}). Realistic depth and soft grounded shadows are achieved through a softbox lighting technique, employing 50 jittered light sources to produce a professional, CAD-like aesthetic. For multi-view grayscale renders, the STL models are rendered from four distinct isometric orientations using this identical setup, and the resulting renders are stitched together into a single composite image.

\paragraph{Photorealistic/Physically-Based Rendering (PBR):} The physically-based renders are produced using \texttt{Pyrender}. We utilize source STEP files for B, F, E, and A, and mesh files for M and O. Material properties for each component are randomly assigned: RGB components are sampled from $U(0, 1)$, while metallicity and roughness are sampled from $U(0.05, 0.95)$. Ground planes are assigned random textures from a curated library of realistic materials (e.g., asphalt, grass, brick, and concrete). Illumination is provided by a three-point lighting configuration (key, fill, and back lights). The camera pose is configured to render an isometric view under perspective projection with a 60° Field of View (FOV). Additionally, the position of the camera is adjusted based on the diameter of the bounding sphere of each model to ensure that the object is not cropped.

We show samples from various \benchmark\ subcategories, along with their corresponding mesh and image modalities, in Figure \ref{fig:samples_modality}.

\begin{figure}[H]
    \centering
    \includegraphics[width=\textwidth]{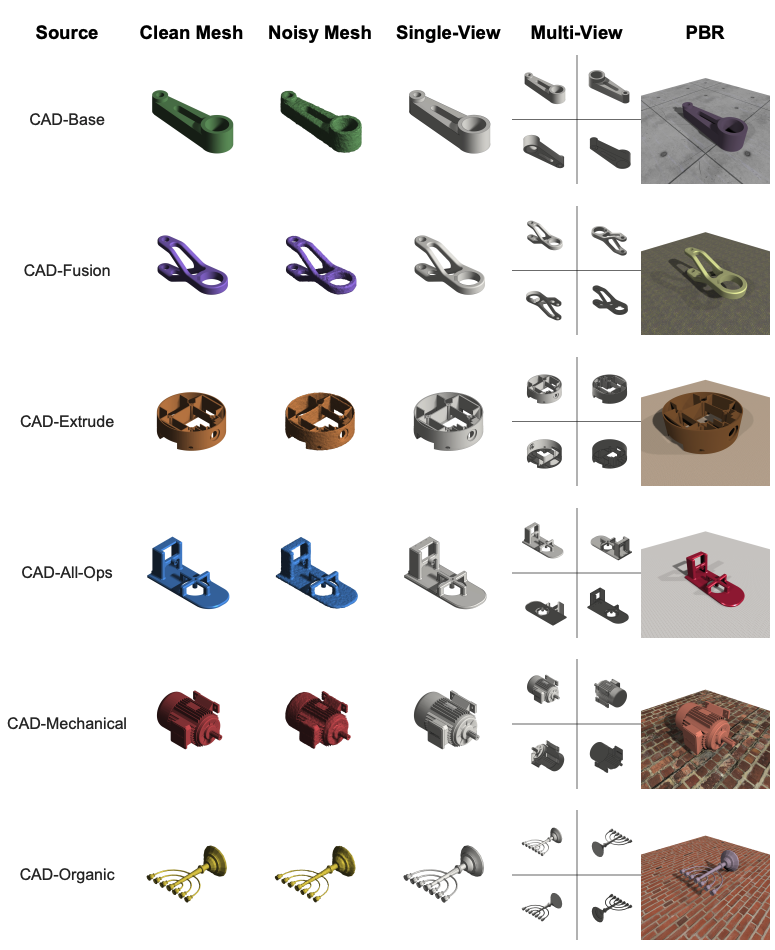}
    \caption{Samples from each subcategory in \benchmark illustrating the available data modalities. From left to right: original clean meshes, their corresponding noisy counterparts, single-view grayscale renders, multi-view grayscale renders, and physically-based renders (PBR) featuring randomized material properties and environmental textures.}
    \label{fig:samples_modality}
\end{figure}

\subsection{More Details on Metrics}
\label{app:metric_details}
We compare the predicted geometry $\hat{S}$ with the ground-truth $S$.

\textbf{Volumetric IoU.}
Let $V(\cdot)$ denote voxelized occupancy. The intersection-over-union is:
\[
\text{IoU} = \frac{|V(\hat{S} \cap S)|}{|V(\hat{S}) \cup V(S)|}
\]

\textbf{Chamfer Distance (CD).}
Let $\hat{P}, P \subset \mathbb{R}^3$ be point samples from $\hat{S}$ and $S$. The symmetric Chamfer distance is:
\[
\text{CD}(\hat{P}, P) = \frac{1}{|\hat{P}|} \sum_{x \in \hat{P}} \min_{y \in P} \|x - y\|_2^2
+ \frac{1}{|P|} \sum_{y \in P} \min_{x \in \hat{P}} \|y - x\|_2^2
\]

\textbf{Surface IoU (SIoU).}
We measure surface alignment using thresholded point-wise coverage. Let $\hat{P}, P$ be point samples from $\hat{S}$ and $S$, and $\tau$ a distance threshold set to $1\%$ of the bounding box diagonal. Then:
\[
\text{SIoU} = \frac{1}{2} \left(
\frac{1}{|\hat{P}|} \sum_{x \in \hat{P}} \mathbb{I}[\min_{y \in P} \|x - y\|_2 < \tau]
+
\frac{1}{|P|} \sum_{y \in P} \mathbb{I}[\min_{x \in \hat{P}} \|y - x\|_2 < \tau]
\right)
\]

\paragraph{Alignment Procedure.}
Before the metrics are measured, we perform an alignment based on the continuous Procrustes analysis solution of aligning two solids~\cite{doris2026cad}. In this way, we do not simply align bounding boxes like many have before~\cite{yu2025gencad,alam2024gencad,kolodiazhnyi2025cadrille,elistratov2026cadevolve,rukhovich2025cad}, but rather perform a mathematically sound affine transformation that includes aligning principal axes. This has been noted as an important step, especially in frontier models where the direction models choose to build geometry along can vary widely, as shown by \citet{doris2026cad}. Overall, the transformation applied to the mesh of the generated code, $\Omega_{g}$, with respect to the ground truth solid $\Omega_{t}$ can be described as:

\[
    \hat{\Omega}_{g} = \{\frac{\mathbf{R}^\star\left(\mathbf{x-\bar{\mathbf{x}}_{g}}\right)+ \bar{\mathbf{x}}_{t}}{\sqrt{\frac{\operatorname{tr}(\mathbf{I_{g}})} {2\times \mathrm{V o l} ( \Omega_{g} )}}}\times\sqrt{\frac{\operatorname{tr}(\mathbf{I_{t}})} {2\times \mathrm{V o l} ( \Omega_{t} )}}\mid \mathbf{x}\in\Omega_{g}\},
\]

where $\bar{\mathbf{x}}$, refers to the centroid, $\mathbf{I}$ refers to the matrix of inertia, and $\mathrm{V o l}$ refers to the volume of solids. Moreover, $\mathbf{R}^\star$ refers to the optimal rotation aligning the principal axes of the two solids determined exhaustively by measuring IoU for all possible 4 (4 $SO(3)$ of 8 possible) ways to align principal axes and picking the transformation with the highest IoU. Note that, besides this, to keep CD numbers consistent with prior works, all ground truth meshes are scaled to have at most a bounding box of largest length $1.0$ as done in prior works.

\paragraph{Executability.}
A predicted program may fail due to invalid operations or inconsistent constraints.

\textbf{Invalid Ratio (IR).}
Let $\mathcal{D}$ be the evaluation set and $\mathbb{I}[\cdot]$ an indicator function:
\[
\text{IR} = \frac{1}{|\mathcal{D}|} \sum_{x \in \mathcal{D}} \mathbb{I}\left[\mathcal{E}(f_\theta(x)) \text{ fails} \right].
\]

We also report \textbf{Valid Shape Rate (VSR)}:
\[
\text{VSR} = 1 - \text{IR}.
\]

\subsection{Model Inference Settings}
\label{app:methods_inference}

\paragraph{Qwen 3.5 9B}
In running Qwen 3.5 9B, we follow the recommended inference settings for instruct (non-thinking) mode on general tasks\footnote{https://huggingface.co/Qwen/Qwen3.5-9B}. Specifically, we disable thinking and use temperature $0.7$, top-$p$ $0.8$, and top-$k$ $20$, with a presence penalty of $1.5$ and repetition penalty of $1.0$. We additionally set $\texttt{min\_p}=0.0$ and a maximum generation length of $4096$ tokens.

\paragraph{Qwen 3.5 27B}
For Qwen 3.5 27B, we use the recommended inference settings for instruct (non-thinking) mode on general tasks\footnote{https://huggingface.co/Qwen/Qwen3.5-27B}. Specifically, we disable thinking and use temperature $0.7$, top-$p$ $0.8$, and top-$k$ $20$, with a presence penalty of $1.5$ and repetition penalty of $1.0$. We additionally set $\texttt{min\_p}=0.0$ and a maximum generation length of $4096$ tokens. These fixed decoding parameters are used across all evaluated splits and input modalities.

\paragraph{Kimi K2.6}
For kimi K 2.6 we use \texttt{kimi-k2.6} through our batch inference script with the default parameters of the API as described in \href{https://platform.moonshot.ai/}{https://platform.moonshot.ai/} with chain of thought (thinking) disabled. Each input includes the rendered CAD image and a fixed CAD-specific prompt requiring executable CadQuery code, all necessary imports, valid solid geometry, parametric dimensions where appropriate, and a final variable named \texttt{result}. The model is instructed to return only Python code, with no explanation.

\paragraph{Gemini 3.1 Pro}
For Gemini 3.1 Pro, we use \texttt{gemini-3.1-pro-preview} through our batch inference script with a maximum generation length of $4096$ tokens. Each input includes the rendered CAD image and a fixed CAD-specific prompt requiring executable CadQuery code, all necessary imports, valid solid geometry, parametric dimensions where appropriate, and a final variable named \texttt{result}. The model is instructed to return only Python code, with no explanation.

\paragraph{GPT 5.4}
In running GPT 5.4, we use the default \texttt{gpt-5.4} model through the OpenAI Batch API with a 24-hour completion window. Each example includes the rendered CAD image at high image detail and a fixed CAD-specific prompt requiring executable CadQuery code, all necessary imports, valid solid geometry, and a final variable named \texttt{result}. The model is instructed to return only Python code, with no explanation. We do not explicitly set the temperature, so the API default is used.

\paragraph{Claude Opus 4.7}
For Claude Opus 4.7, we use \texttt{claude-opus-4-7} with a maximum generation length of $4096$ tokens. Each query includes the rendered CAD image and the same CAD-specific prompt used for the other proprietary vision-language models, requiring executable CadQuery code, all necessary imports, valid solid geometry, and a final variable named \texttt{result}. The model is instructed to return only Python code, with no explanation.

\section{Expanded Comparison of CADBench with Other Benchmarks}
\label{app:related_work}
We evaluate \benchmark relative to established benchmarks, such as DeepCAD, Fusion 360 Reconstruction, MCB, and Omni-CAD, across three primary axes: complexity, similarity, and geometric canonicality. We quantify complexity via B-Rep face count. Similarity of objects within a benchmark is assessed using the pairwise cosine similarities between the feature vectors of all objects in the set (see Appendix \ref{app:dino_embeddings} for the procedure to get the feature vectors). To further characterize geometric canonicality, we measure the similarity of benchmark objects to a unit cube using two metrics: (i) the volumetric Intersection over Union (IoU) and (ii) the cosine similarity between the feature vector of the object and that of a unit cube.

\begin{table}[H]
    \centering
    \small
    \setlength{\tabcolsep}{3.5pt}
    \renewcommand{\arraystretch}{1.3}
    \caption{Comparison of structural complexity, diversity, and canonicality across benchmarks. \textbf{Complexity} is measured by B-Rep face count ($\uparrow$). \textbf{Similarity} measures dataset redundancy ($\downarrow$ indicates higher diversity). \textbf{Geometric Canonicality} assesses proximity to a unit cube ($\downarrow$ indicates models further from simple primitives).}
    \label{tab:benchmark_comparison}
    \resizebox{\textwidth}{!}{%
    \begin{tabular}{ll cccc cc cc cc}
        \toprule
        \multirow{3}{*}{\textbf{Bench.}} & \multirow{3}{*}{\textbf{Split}} & \multicolumn{4}{c}{\textbf{Complexity} $\uparrow$} & \multicolumn{2}{c}{\textbf{Similarity} $\downarrow$} & \multicolumn{4}{c}{\textbf{Geometric Canonicality} $\downarrow$} \\
        \cmidrule(lr){3-6} \cmidrule(lr){7-8} \cmidrule(lr){9-12}
        & & \multirow{2}{*}{Mean} & \multirow{2}{*}{Med.} & \multirow{2}{*}{Min.} & \multirow{2}{*}{Max.} & \multirow{2}{*}{Mean} & \multirow{2}{*}{Med.} & \multicolumn{2}{c}{IoU-based} & \multicolumn{2}{c}{DINO-based} \\
        \cmidrule(lr){9-10} \cmidrule(lr){11-12}
        & & & & & & & & Mean & Med. & Mean & Med. \\
        \midrule
        \multirow{3}{*}{\textbf{B}} & Easy   & 6.55 & 7 & 3 & 9 & 0.514 & 0.499 & 0.210 & 0.143 & 0.489 & 0.449 \\
                                    & Medium & 12.93 & 12 & 10 & 18 & 0.512 & 0.509 & 0.177 & 0.104 & 0.490 & 0.459 \\
                                    & Hard   & 30.23 & 27 & 19 & 110 & 0.506 & 0.500 & 0.180 & 0.129 & 0.477 & 0.438 \\
        \midrule
        \multirow{3}{*}{\textbf{F}} & Easy   & 5.08 & 5 & 1 & 7 & 0.451 & 0.435 & 0.187 & 0.110 & 0.395 & 0.364 \\
                                    & Medium & 16.27 & 13 & 8 & 55 & 0.434 & 0.423 & 0.153 & 0.093 & 0.363 & 0.327 \\
                                    & Hard   & 93.32 & 79 & 57 & 421 & 0.375 & 0.360 & 0.146 & 0.086 & 0.297 & 0.276 \\
        \midrule
        \multirow{3}{*}{\textbf{E}} & Easy   & 11.26 & 10 & 3 & 54 & 0.470 & 0.457 & 0.164 & 0.095 & 0.461 & 0.426 \\
                                    & Medium & 91.43 & 80 & 57 & 179 & 0.358 & 0.342 & 0.090 & 0.044 & 0.334 & 0.308 \\
                                    & Hard   & 687.23 & 358 & 181 & 19851 & 0.373 & 0.341 & 0.088 & 0.042 & 0.370 & 0.327 \\
        \midrule
        \multirow{3}{*}{\textbf{A}} & Easy   & 13.81 & 12 & 2 & 36 & 0.454 & 0.445 & 0.198 & 0.141 & 0.371 & 0.337 \\
                                    & Medium & 59.59 & 51 & 37 & 137 & 0.396 & 0.384 & 0.137 & 0.080 & 0.301 & 0.280 \\
                                    & Hard   & 383.71 & 206 & 138 & 12292 & 0.346 & 0.323 & 0.131 & 0.074 & 0.288 & 0.255 \\
        \midrule
        \textbf{M}                  & ---    & --- & --- & --- & --- & 0.464 & 0.448 & 0.177 & 0.115 & 0.351 & 0.353 \\
        \textbf{O}                  & ---    & --- & --- & --- & --- & 0.306 & 0.288 & 0.196 & 0.118 & 0.294 & 0.262 \\
        \midrule
        \textbf{Aggregate}$^\ddagger$ & ---    & \textbf{117.62} & --- & 1 & 19851 & \textbf{0.417} & --- & 0.166 & --- & 0.365 & --- \\
        \specialrule{.12em}{.05em}{.05em}
        \textbf{DeepCAD}            & ---    & 12.94 & 9 & 3 & 110 & 0.533 & 0.511 & 0.229 & 0.151 & 0.561 & 0.517 \\
        \textbf{Fusion 360}         & ---    & 16.25 & 10 & 3 & 388 & 0.456 & 0.440 & 0.165 & 0.101 & 0.409 & 0.363 \\
        \textbf{MCB}                & ---    & --- & --- & --- & --- & 0.503 & 0.491 & \textbf{0.156} & 0.090 & \textbf{0.301} & 0.296 \\
        \textbf{Omni-CAD}           & ---    & 25.97 & 11 & 2 & 5448 & 0.483 & 0.464 & 0.189 & 0.104 & 0.503 & 0.450 \\
        \bottomrule
    \end{tabular}}
    \vspace{2pt}
    {\raggedright \footnotesize $^\ddagger$ Values represent a weighted average across all subsets. $\uparrow$ indicates increasing complexity; $\downarrow$ indicates lower similarity (corresponding to higher diversity or lower canonicality).\par}
\end{table}

\section{Additional Results}
\label{app:results}
\subsection{Complete Benchmark Results Given Ideal Inputs}
\subsubsection{Per-Split Results}
\label{app:full_results_ideal_input}
Complete benchmark results given ideal inputs -- singleview gray image and mesh -- can be seen in Table \ref{tab:complete_fidelity_metrics} (geometric fidelity metrics), Table \ref{tab:complete_validity_metrics} (executability metrics), and Table \ref{tab:complete_quality_metrics} (program quality metrics). Similar complete tables could be created for the other modality inputs. Data for these results can be found on our GitHub repo.

\begin{table}[H]
\centering
\small
\setlength{\tabcolsep}{3.2pt}
\renewcommand{\arraystretch}{1.08}
\caption{\textbf{Geometric fidelity metrics} for models tested on ideal inputs.}
\label{tab:complete_fidelity_metrics}
\resizebox{\textwidth}{!}{%
\begin{tabular}{llccccccccccccccc}
\toprule
& & \multicolumn{4}{c}{Low Face Count} & \multicolumn{4}{c}{Medium Face Count} & \multicolumn{4}{c}{High Face Count} & \multicolumn{2}{c}{Other} & \\
\cmidrule(lr){3-6} \cmidrule(lr){7-10} \cmidrule(lr){11-14} \cmidrule(lr){15-16}
Metric & Model 
& B & F & E & A 
& B & F & E & A 
& B & F & E & A 
& M & O
& \textbf{Aggregate} \\
\midrule
\multirow{11}{*}{\textbf{IoU}}
& CADFit
& \heatcell{0.986} & \heatcell{0.983} & \heatcell{0.973} & \heatcell{0.934} & \heatcell{0.966} & \heatcell{0.955} & \heatcell{0.835} & \heatcell{0.789} & \heatcell{0.919} & \heatcell{0.868} & \heatcell{0.835} & \heatcell{0.789} & \heatcell{0.915} & \heatcell{0.625} & \textbf{\heatcell{0.859}} \\
& CAD-Recode
& \heatcell[0.001]{0.963} & \heatcell[0.002]{0.917} & \heatcell[0.002]{0.917} & \heatcell[0.007]{0.837} & \heatcell[0.003]{0.908} & \heatcell[0.006]{0.816} & \heatcell[0.016]{0.424} & \heatcell[0.015]{0.485} & \heatcell[0.003]{0.804} & \heatcell[0.010]{0.546} & \heatcell[0.016]{0.540} & \heatcell[0.008]{0.415} & \heatcell[0.006]{0.775} & \heatcell[0.014]{0.409} & \textbf{\heatcell{0.673}} \\
& CADEvolve
& \heatcell[0.000]{0.970} & \heatcell[0.002]{0.917} & \heatcell[0.004]{0.928} & \heatcell[0.002]{0.855} & \heatcell[0.002]{0.924} & \heatcell[0.000]{0.827} & \heatcell[0.003]{0.548} & \heatcell[0.012]{0.541} & \heatcell[0.003]{0.824} & \heatcell[0.003]{0.605} & \heatcell[0.005]{0.621} & \heatcell[0.022]{0.480} & \heatcell[0.006]{0.792} & \heatcell[0.004]{0.439} & \textbf{\heatcell{0.707}} \\
& Cadrille
& \heatcell[0.000]{0.961} & \heatcell[0.002]{0.911} & \heatcell[0.003]{0.911} & \heatcell[0.007]{0.840} & \heatcell[0.003]{0.904} & \heatcell[0.004]{0.813} & \heatcell[0.008]{0.485} & \heatcell[0.007]{0.514} & \heatcell[0.003]{0.808} & \heatcell[0.033]{0.584} & \heatcell[0.002]{0.554} & \heatcell[0.021]{0.430} & \heatcell[0.014]{0.783} & \heatcell[0.002]{0.434} & \textbf{\heatcell{0.687}} \\
\cmidrule(lr){2-17}
& Claude Opus
& \heatcell[0.003]{0.642} & \heatcell[0.010]{0.582} & \heatcell[0.003]{0.570} & \heatcell[0.007]{0.524} & \heatcell[0.001]{0.547} & \heatcell[0.004]{0.432} & \heatcell[0.012]{0.290} & \heatcell[0.005]{0.295} & \heatcell[0.002]{0.444} & \heatcell[0.013]{0.318} & \heatcell[0.009]{0.312} & \heatcell[0.004]{0.275} & \heatcell[0.002]{0.505} & \heatcell[0.009]{0.221} & \textbf{\heatcell{0.412}} \\
& Gemini
& \heatcell[0.007]{0.705} & \heatcell[0.005]{0.637} & \heatcell[0.010]{0.612} & \heatcell[0.010]{0.554} & \heatcell[0.004]{0.588} & \heatcell[0.007]{0.473} & \heatcell[0.014]{0.282} & \heatcell[0.024]{0.210} & \heatcell[0.004]{0.476} & \heatcell[0.034]{0.252} & \heatcell[0.021]{0.251} & \heatcell[0.044]{0.045} & \heatcell[0.008]{0.493} & \heatcell[0.011]{0.104} & \textbf{\heatcell{0.382}} \\
& GPT 5.4
& \heatcell[0.007]{0.618} & \heatcell[0.006]{0.493} & \heatcell[0.027]{0.488} & \heatcell[0.021]{0.226} & \heatcell[0.015]{0.474} & \heatcell[0.004]{0.235} & \heatcell[0.000]{0.000} & \heatcell[0.000]{0.000} & \heatcell[0.009]{0.307} & \heatcell[0.000]{0.000} & \heatcell[0.000]{0.000} & \heatcell[0.000]{0.000} & \heatcell[0.000]{0.000} & \heatcell[0.000]{0.000} & \textbf{\heatcell{0.158}} \\
& Kimi K2.6
&  \heatcell[0.006]{0.587} & \heatcell[0.002]{0.479} & \heatcell[0.005]{0.487} & \heatcell[0.013]{0.408} & \heatcell[0.005]{0.444} & \heatcell[0.010]{0.272} & \heatcell[0.019]{0.110} & \heatcell[0.012]{0.046} & \heatcell[0.001]{0.344} & \heatcell[0.008]{0.032} & \heatcell[0.039]{0.043} & \heatcell[0.000]{0.000} & \heatcell[0.019]{0.260} & \heatcell[0.000]{0.000} & \textbf{\heatcell{0.224}} \\
& Qwen 27B
& \heatcell[0.021]{0.226} & \heatcell[0.015]{0.026} & \heatcell[0.016]{0.017} & \heatcell[0.000]{0.000} & \heatcell[0.000]{0.000} & \heatcell[0.000]{0.000} & \heatcell[0.000]{0.000} & \heatcell[0.000]{0.000} & \heatcell[0.000]{0.000} & \heatcell[0.000]{0.000} & \heatcell[0.000]{0.000} & \heatcell[0.000]{0.000} & \heatcell[0.000]{0.000} & \heatcell[0.000]{0.000} & \textbf{\heatcell{0.015}} \\
& Qwen 9B
& \heatcell[0.000]{0.000} & \heatcell[0.000]{0.000} & \heatcell[0.000]{0.000} & \heatcell[0.000]{0.000} & \heatcell[0.000]{0.000} & \heatcell[0.000]{0.000} & \heatcell[0.000]{0.000} & \heatcell[0.000]{0.000} & \heatcell[0.000]{0.000} & \heatcell[0.000]{0.000} & \heatcell[0.000]{0.000} & \heatcell[0.000]{0.000} & \heatcell[0.000]{0.000} & \heatcell[0.000]{0.000} & \textbf{\heatcell{0.000}} \\
& CADCoder
& \heatcell{0.664} & \heatcell{0.563} & \heatcell{0.545} & \heatcell{0.429} & \heatcell{0.479} & \heatcell{0.387} & \heatcell{0.194} & \heatcell{0.216} & \heatcell{0.351} & \heatcell{0.215} & \heatcell{0.274} & \heatcell{0.198} & \heatcell{0.433} & \heatcell{0.187} & \textbf{\heatcell{0.354}} \\
\midrule
\multirow{11}{*}{\textbf{CD}}
& CADFit
& \heatcellcd{0.027} & \heatcellcd{0.025} & \heatcellcd{0.026} & \heatcellcd{0.033} & \heatcellcd{0.028} & \heatcellcd{0.027} & \heatcellcd{0.032} & \heatcellcd{0.041} & \heatcellcd{0.033} & \heatcellcd{0.035} & \heatcellcd{0.031} & \heatcellcd{0.041} & \heatcellcd{0.034} & \heatcellcd{0.066} & \textbf{\heatcellcd{0.038}} \\
& CAD-Recode
& \heatcellcd[0.000]{0.028} & \heatcellcd[0.000]{0.029} & \heatcellcd[0.000]{0.030} & \heatcellcd[0.001]{0.040} & \heatcellcd[0.000]{0.032} & \heatcellcd[0.001]{0.036} & \heatcellcd[0.000]{0.060} & \heatcellcd[0.003]{0.073} & \heatcellcd[0.000]{0.041} & \heatcellcd[0.002]{0.061} & \heatcellcd[0.001]{0.041} & \heatcellcd[0.002]{0.072} & \heatcellcd[0.001]{0.046} & \heatcellcd[0.002]{0.109} & \textbf{\heatcellcd{0.056}} \\
& CADEvolve
& \heatcellcd[0.000]{0.030} & \heatcellcd[0.000]{0.032} & \heatcellcd[0.000]{0.031} & \heatcellcd[0.001]{0.041} & \heatcellcd[0.000]{0.034} & \heatcellcd[0.000]{0.040} & \heatcellcd[0.001]{0.062} & \heatcellcd[0.001]{0.072} & \heatcellcd[0.000]{0.044} & \heatcellcd[0.001]{0.065} & \heatcellcd[0.000]{0.049} & \heatcellcd[0.001]{0.075} & \heatcellcd[0.001]{0.051} & \heatcellcd[0.002]{0.129} & \textbf{\heatcellcd{0.062}} \\
& Cadrille
& \heatcellcd[0.000]{0.028} & \heatcellcd[0.000]{0.030} & \heatcellcd[0.000]{0.030} & \heatcellcd[0.001]{0.040} & \heatcellcd[0.000]{0.032} & \heatcellcd[0.000]{0.038} & \heatcellcd[0.001]{0.057} & \heatcellcd[0.004]{0.074} & \heatcellcd[0.000]{0.042} & \heatcellcd[0.003]{0.060} & \heatcellcd[0.001]{0.044} & \heatcellcd[0.001]{0.073} & \heatcellcd[0.003]{0.047} & \heatcellcd[0.000]{0.104} & \textbf{\heatcellcd{0.056}} \\
\cmidrule(lr){2-17}
& Claude Opus
& \heatcellcd[0.001]{0.077} & \heatcellcd[0.001]{0.080} & \heatcellcd[0.002]{0.075} & \heatcellcd[0.005]{0.102} & \heatcellcd[0.002]{0.093} & \heatcellcd[0.003]{0.103} & \heatcellcd[0.002]{0.078} & \heatcellcd[0.001]{0.118} & \heatcellcd[0.002]{0.119} & \heatcellcd[0.001]{0.103} & \heatcellcd[0.002]{0.052} & \heatcellcd[0.002]{0.093} & \heatcellcd[0.001]{0.095} & \heatcellcd[0.003]{0.146} & \textbf{\heatcellcd{0.101}} \\
& Gemini
& \heatcellcd[0.001]{0.061} & \heatcellcd[0.001]{0.066} & \heatcellcd[0.001]{0.066} & \heatcellcd[0.003]{0.087} & \heatcellcd[0.001]{0.080} & \heatcellcd[0.001]{0.092} & \heatcellcd[0.001]{0.070} & \heatcellcd[0.005]{0.099} & \heatcellcd[0.002]{0.104} & \heatcellcd[0.003]{0.090} & \heatcellcd[0.002]{0.052} & \heatcellcd[0.002]{0.078} & \heatcellcd[0.001]{0.081} & \heatcellcd[0.001]{0.126} & \textbf{\heatcellcd{0.087}} \\
& GPT 5.4
& \heatcellcd[0.001]{0.087} & \heatcellcd[0.003]{0.088} & \heatcellcd[0.003]{0.089} & \heatcellcd[0.004]{0.110} & \heatcellcd[0.003]{0.109} & \heatcellcd[0.005]{0.119} & \heatcellcd[0.002]{0.086} & \heatcellcd[0.004]{0.132} & \heatcellcd[0.003]{0.142} & \heatcellcd[0.006]{0.121} & \heatcellcd[0.003]{0.057} & \heatcellcd[0.003]{0.089} & \heatcellcd[0.000]{0.086} & \heatcellcd[0.005]{0.155} & \textbf{\heatcellcd{0.109}} \\
& Kimi K2.6
& \heatcellcd[0.002]{0.083} & \heatcellcd[0.002]{0.085} & \heatcellcd[0.004]{0.091} & \heatcellcd[0.008]{0.108} & \heatcellcd[0.005]{0.116} & \heatcellcd[0.005]{0.120} & \heatcellcd[0.004]{0.097} & \heatcellcd[0.003]{0.133} & \heatcellcd[0.005]{0.145} & \heatcellcd[0.006]{0.120} & \heatcellcd[0.002]{0.055} & \heatcellcd[0.005]{0.101} & \heatcellcd[0.001]{0.092} & \heatcellcd[0.003]{0.158} & \textbf{\heatcellcd{0.111}} \\
& Qwen 27B
& \heatcellcd[0.007]{0.161} & \heatcellcd[0.004]{0.155} & \heatcellcd[0.002]{0.158} & \heatcellcd[0.009]{0.177} & \heatcellcd[0.003]{0.182} & \heatcellcd[0.004]{0.188} & \heatcellcd[0.009]{0.155} & \heatcellcd[0.011]{0.194} & \heatcellcd[0.004]{0.211} & \heatcellcd[0.014]{0.191} & \heatcellcd[0.004]{0.119} & \heatcellcd[0.005]{0.157} & \heatcellcd[0.004]{0.154} & \heatcellcd[0.004]{0.220} & \textbf{\heatcellcd{0.176}} \\
& Qwen 9B
& \heatcellcd[0.002]{0.170} & \heatcellcd[0.009]{0.139} & \heatcellcd[0.003]{0.168} & \heatcellcd[0.010]{0.210} & \heatcellcd[0.019]{0.204} & \heatcellcd[0.008]{0.213} & \heatcellcd[0.041]{0.175} & \heatcellcd[0.027]{0.206} & \heatcellcd[0.009]{0.222} & \heatcellcd[0.024]{0.199} & \heatcellcd[0.027]{0.112} & \heatcellcd[0.061]{0.168} & \heatcellcd[0.010]{0.209} & \heatcellcd[0.022]{0.232} & \textbf{\heatcellcd{0.195}} \\
& CADCoder
& \heatcellcd{0.078} & \heatcellcd{0.092} & \heatcellcd{0.097} & \heatcellcd{0.154} & \heatcellcd{0.133} & \heatcellcd{0.154} & \heatcellcd{0.151} & \heatcellcd{0.194} & \heatcellcd{0.171} & \heatcellcd{0.186} & \heatcellcd{0.115} & \heatcellcd{0.173} & \heatcellcd{0.140} & \heatcellcd{0.237} & \textbf{\heatcellcd{0.157}} \\
\midrule
\multirow{11}{*}{\textbf{SIoU}}
& CADFit
& \heatcell{0.833} & \heatcell{0.873} & \heatcell{0.842} & \heatcell{0.710} & \heatcell{0.812} & \heatcell{0.823} & \heatcell{0.761} & \heatcell{0.609} & \heatcell{0.718} & \heatcell{0.708} & \heatcell{0.744} & \heatcell{0.605} & \heatcell{0.697} & \heatcell{0.366} & \textbf{\heatcell{0.679}} \\
& CAD-Recode
& \heatcell[0.004]{0.806} & \heatcell[0.002]{0.778} & \heatcell[0.003]{0.771} & \heatcell[0.005]{0.567} & \heatcell[0.005]{0.734} & \heatcell[0.021]{0.650} & \heatcell[0.013]{0.455} & \heatcell[0.024]{0.318} & \heatcell[0.009]{0.586} & \heatcell[0.014]{0.402} & \heatcell[0.014]{0.590} & \heatcell[0.004]{0.328} & \heatcell[0.005]{0.519} & \heatcell[0.006]{0.173} & \textbf{\heatcell{0.503}} \\
& CADEvolve
& \heatcell[0.005]{0.793} & \heatcell[0.005]{0.769} & \heatcell[0.006]{0.774} & \heatcell[0.011]{0.588} & \heatcell[0.005]{0.729} & \heatcell[0.009]{0.639} & \heatcell[0.004]{0.493} & \heatcell[0.006]{0.302} & \heatcell[0.008]{0.590} & \heatcell[0.009]{0.403} & \heatcell[0.007]{0.580} & \heatcell[0.012]{0.290} & \heatcell[0.004]{0.520} & \heatcell[0.003]{0.154} & \textbf{\heatcell{0.498}} \\
& Cadrille
& \heatcell[0.001]{0.804} & \heatcell[0.003]{0.772} & \heatcell[0.004]{0.767} & \heatcell[0.009]{0.587} & \heatcell[0.006]{0.720} & \heatcell[0.005]{0.638} & \heatcell[0.015]{0.508} & \heatcell[0.007]{0.350} & \heatcell[0.006]{0.583} & \heatcell[0.038]{0.422} & \heatcell[0.001]{0.623} & \heatcell[0.002]{0.347} & \heatcell[0.016]{0.536} & \heatcell[0.002]{0.193} & \textbf{\heatcell{0.517}} \\
\cmidrule(lr){2-17}
& Claude Opus
& \heatcell[0.005]{0.168} & \heatcell[0.003]{0.144} & \heatcell[0.009]{0.190} & \heatcell[0.003]{0.111} & \heatcell[0.005]{0.160} & \heatcell[0.001]{0.111} & \heatcell[0.011]{0.142} & \heatcell[0.002]{0.091} & \heatcell[0.004]{0.110} & \heatcell[0.005]{0.092} & \heatcell[0.002]{0.103} & \heatcell[0.005]{0.084} & \heatcell[0.004]{0.077} & \heatcell[0.002]{0.068} & \textbf{\heatcell{0.108}} \\
& Gemini
& \heatcell[0.010]{0.252} & \heatcell[0.010]{0.210} & \heatcell[0.010]{0.226} & \heatcell[0.006]{0.120} & \heatcell[0.011]{0.188} & \heatcell[0.007]{0.118} & \heatcell[0.008]{0.117} & \heatcell[0.006]{0.057} & \heatcell[0.007]{0.113} & \heatcell[0.005]{0.064} & \heatcell[0.010]{0.079} & \heatcell[0.015]{0.016} & \heatcell[0.004]{0.074} & \heatcell[0.002]{0.041} & \textbf{\heatcell{0.106}} \\
& GPT 5.4
& \heatcell[0.005]{0.119} & \heatcell[0.002]{0.073} & \heatcell[0.016]{0.086} & \heatcell[0.003]{0.033} & \heatcell[0.003]{0.090} & \heatcell[0.006]{0.051} & \heatcell[0.000]{0.000} & \heatcell[0.000]{0.000} & \heatcell[0.005]{0.065} & \heatcell[0.000]{0.000} & \heatcell[0.000]{0.000} & \heatcell[0.000]{0.000} & \heatcell[0.000]{0.000} & \heatcell[0.000]{0.000} & \textbf{\heatcell{0.029}} \\
& Kimi K2.6
& \heatcell[0.008]{0.113} & \heatcell[0.009]{0.075} & \heatcell[0.006]{0.100} & \heatcell[0.003]{0.058} & \heatcell[0.003]{0.082} & \heatcell[0.002]{0.051} & \heatcell[0.004]{0.050} & \heatcell[0.005]{0.025} & \heatcell[0.004]{0.071} & \heatcell[0.001]{0.016} & \heatcell[0.016]{0.018} & \heatcell[0.000]{0.000} & \heatcell[0.000]{0.031} & \heatcell[0.000]{0.000} & \textbf{\heatcell{0.042}} \\
& Qwen 27B
& \heatcell[0.001]{0.023} & \heatcell[0.009]{0.009} & \heatcell[0.004]{0.002} & \heatcell[0.000]{0.000} & \heatcell[0.000]{0.000} & \heatcell[0.000]{0.000} & \heatcell[0.000]{0.000} & \heatcell[0.000]{0.000} & \heatcell[0.000]{0.000} & \heatcell[0.000]{0.000} & \heatcell[0.000]{0.000} & \heatcell[0.000]{0.000} & \heatcell[0.000]{0.000} & \heatcell[0.000]{0.000} & \textbf{\heatcell{0.002}} \\
& Qwen 9B
& \heatcell[0.000]{0.000} & \heatcell[0.000]{0.000} & \heatcell[0.000]{0.000} & \heatcell[0.000]{0.000} & \heatcell[0.000]{0.000} & \heatcell[0.000]{0.000} & \heatcell[0.000]{0.000} & \heatcell[0.000]{0.000} & \heatcell[0.000]{0.000} & \heatcell[0.000]{0.000} & \heatcell[0.000]{0.000} & \heatcell[0.000]{0.000} & \heatcell[0.000]{0.000} & \heatcell[0.000]{0.000} & \textbf{\heatcell{0.000}} \\
& CADCoder
& \heatcell{0.197} & \heatcell{0.193} & \heatcell{0.201} & \heatcell{0.094} & \heatcell{0.132} & \heatcell{0.108} & \heatcell{0.140} & \heatcell{0.091} & \heatcell{0.086} & \heatcell{0.086} & \heatcell{0.213} & \heatcell{0.100} & \heatcell{0.080} & \heatcell{0.064} & \textbf{\heatcell{0.115}} \\
\bottomrule
\end{tabular}%
}
\end{table}

\begin{table}[H]
\centering
\small
\setlength{\tabcolsep}{3.2pt}
\renewcommand{\arraystretch}{1.08}
\caption{\textbf{VSR} for models tested on ideal inputs.}
\label{tab:complete_validity_metrics}
\resizebox{\textwidth}{!}{%
\begin{tabular}{llccccccccccccccc}
\toprule
& & \multicolumn{4}{c}{Low Face Count} & \multicolumn{4}{c}{Medium Face Count} & \multicolumn{4}{c}{High Face Count} & \multicolumn{2}{c}{Other} & \\
\cmidrule(lr){3-6} \cmidrule(lr){7-10} \cmidrule(lr){11-14} \cmidrule(lr){15-16}
Metric & Model 
& B & F & E & A 
& B & F & E & A 
& B & F & E & A 
& M & O
& \textbf{Aggregate} \\
\midrule
\multirow{11}{*}{\textbf{VSR}}
& CADFit
& \heatcell{0.981} & \heatcell{0.985} & \heatcell{0.985} & \heatcell{0.988} & \heatcell{0.990} & \heatcell{0.995} & \heatcell{0.989} & \heatcell{0.991} & \heatcell{0.996} & \heatcell{0.993} & \heatcell{0.929} & \heatcell{0.980} & \heatcell{0.994} & \heatcell{0.956} & \textbf{\heatcell{0.981}} \\
& CAD-Recode
& \heatcell[0.004]{0.960} & \heatcell[0.007]{0.942} & \heatcell[0.005]{0.957} & \heatcell[0.003]{0.927} & \heatcell[0.005]{0.956} & \heatcell[0.009]{0.934} & \heatcell[0.009]{0.864} & \heatcell[0.007]{0.900} & \heatcell[0.006]{0.945} & \heatcell[0.005]{0.884} & \heatcell[0.009]{0.849} & \heatcell[0.007]{0.882} & \heatcell[0.001]{0.931} & \heatcell[0.009]{0.875} & \textbf{\heatcell{0.912}} \\
& CADEvolve
& \heatcell[0.002]{0.986} & \heatcell[0.003]{0.985} & \heatcell[0.001]{0.983} & \heatcell[0.004]{0.969} & \heatcell[0.000]{0.996} & \heatcell[0.006]{0.978} & \heatcell[0.001]{0.983} & \heatcell[0.003]{0.927} & \heatcell[0.002]{0.991} & \heatcell[0.004]{0.948} & \heatcell[0.005]{0.972} & \heatcell[0.008]{0.907} & \heatcell[0.002]{0.963} & \heatcell[0.003]{0.965} & \textbf{\heatcell{0.967}} \\
& Cadrille
& \heatcell[0.003]{0.967} & \heatcell[0.001]{0.960} & \heatcell[0.003]{0.965} & \heatcell[0.001]{0.950} & \heatcell[0.005]{0.961} & \heatcell[0.008]{0.945} & \heatcell[0.001]{0.911} & \heatcell[0.006]{0.928} & \heatcell[0.005]{0.959} & \heatcell[0.017]{0.921} & \heatcell[0.006]{0.913} & \heatcell[0.010]{0.908} & \heatcell[0.010]{0.963} & \heatcell[0.003]{0.907} & \textbf{\heatcell{0.939}} \\
\cmidrule(lr){2-17}
& Claude Opus
& \heatcell[0.004]{0.951} & \heatcell[0.005]{0.909} & \heatcell[0.008]{0.922} & \heatcell[0.020]{0.867} & \heatcell[0.002]{0.919} & \heatcell[0.011]{0.848} & \heatcell[0.012]{0.725} & \heatcell[0.004]{0.742} & \heatcell[0.011]{0.887} & \heatcell[0.010]{0.728} & \heatcell[0.006]{0.631} & \heatcell[0.019]{0.681} & \heatcell[0.009]{0.814} & \heatcell[0.022]{0.703} & \textbf{\heatcell{0.798}} \\
& Gemini
& \heatcell[0.008]{0.957} & \heatcell[0.001]{0.913} & \heatcell[0.010]{0.913} & \heatcell[0.012]{0.824} & \heatcell[0.010]{0.927} & \heatcell[0.003]{0.838} & \heatcell[0.014]{0.660} & \heatcell[0.015]{0.594} & \heatcell[0.006]{0.869} & \heatcell[0.017]{0.613} & \heatcell[0.018]{0.585} & \heatcell[0.015]{0.509} & \heatcell[0.008]{0.732} & \heatcell[0.004]{0.572} & \textbf{\heatcell{0.729}} \\
& GPT 5.4
& \heatcell[0.003]{0.891} & \heatcell[0.009]{0.784} & \heatcell[0.053]{0.786} & \heatcell[0.016]{0.594} & \heatcell[0.014]{0.816} & \heatcell[0.005]{0.655} & \heatcell[0.027]{0.413} & \heatcell[0.010]{0.375} & \heatcell[0.009]{0.742} & \heatcell[0.014]{0.349} & \heatcell[0.021]{0.376} & \heatcell[0.011]{0.287} & \heatcell[0.002]{0.447} & \heatcell[0.010]{0.275} & \textbf{\heatcell{0.513}} \\
& Kimi K2.6
& \heatcell[0.006]{0.900} & \heatcell[0.012]{0.768} & \heatcell[0.008]{0.848} & \heatcell[0.021]{0.702} & \heatcell[0.004]{0.837} & \heatcell[0.013]{0.685} & \heatcell[0.011]{0.588} & \heatcell[0.011]{0.527} & \heatcell[0.005]{0.806} & \heatcell[0.004]{0.516} & \heatcell[0.019]{0.518} & \heatcell[0.029]{0.467} & \heatcell[0.004]{0.572} & \heatcell[0.004]{0.402} & \textbf{\heatcell{0.616}} \\
& Qwen 27B
& \heatcell[0.010]{0.630} & \heatcell[0.016]{0.528} & \heatcell[0.014]{0.505} & \heatcell[0.006]{0.403} & \heatcell[0.017]{0.450} & \heatcell[0.011]{0.366} & \heatcell[0.012]{0.172} & \heatcell[0.013]{0.184} & \heatcell[0.006]{0.393} & \heatcell[0.003]{0.173} & \heatcell[0.006]{0.165} & \heatcell[0.001]{0.131} & \heatcell[0.004]{0.368} & \heatcell[0.002]{0.276} & \textbf{\heatcell{0.335}} \\
& Qwen 9B
& \heatcell[0.021]{0.264} & \heatcell[0.013]{0.185} & \heatcell[0.019]{0.192} & \heatcell[0.012]{0.112} & \heatcell[0.004]{0.174} & \heatcell[0.013]{0.090} & \heatcell[0.004]{0.036} & \heatcell[0.002]{0.029} & \heatcell[0.002]{0.132} & \heatcell[0.005]{0.031} & \heatcell[0.003]{0.044} & \heatcell[0.009]{0.026} & \heatcell[0.009]{0.104} & \heatcell[0.007]{0.053} & \textbf{\heatcell{0.099}} \\
& CADCoder
& \heatcell{0.979} & \heatcell{0.979} & \heatcell{0.985} & \heatcell{0.982} & \heatcell{0.985} & \heatcell{0.979} & \heatcell{0.904} & \heatcell{0.944} & \heatcell{0.973} & \heatcell{0.914} & \heatcell{0.902} & \heatcell{0.878} & \heatcell{0.977} & \heatcell{0.899} & \textbf{\heatcell{0.946}} \\
\bottomrule
\end{tabular}%
}
\end{table}

\begin{table}[H]
\centering
\small
\setlength{\tabcolsep}{3.2pt}
\renewcommand{\arraystretch}{1.08}
\caption{\textbf{Program quality metrics} for models tested on ideal inputs.}
\label{tab:complete_quality_metrics}
\resizebox{\textwidth}{!}{%
\begin{tabular}{llccccccccccccccc}
\toprule
& & \multicolumn{4}{c}{Low Face Count} & \multicolumn{4}{c}{Medium Face Count} & \multicolumn{4}{c}{High Face Count} & \multicolumn{2}{c}{Other} & \\
\cmidrule(lr){3-6} \cmidrule(lr){7-10} \cmidrule(lr){11-14} \cmidrule(lr){15-16}
Metric & Model 
& B & F & E & A 
& B & F & E & A 
& B & F & E & A 
& M & O
& \textbf{Aggregate} \\
\midrule
\multirow{11}{*}{\textbf{Token Count}}
& CADFit
& 154 & 183 & 192 & 266 & 224 & 305 & 804 & 533 & 300 & 779 & 1037 & 918 & 417 & 999 & \textbf{552} \\
& CAD-Recode
& \noheatcell[2]{76} & \noheatcell[1]{86} & \noheatcell[2]{112} & \noheatcell[6]{178} & \noheatcell[0]{144} & \noheatcell[4]{178} & \noheatcell[6]{356} & \noheatcell[3]{346} & \noheatcell[2]{200} & \noheatcell[6]{348} & \noheatcell[5]{282} & \noheatcell[5]{392} & \noheatcell[1]{178} & \noheatcell[4]{337} & \textbf{\noheatcell{236}} \\
& CADEvolve
& \noheatcell[0]{123} & \noheatcell[1]{118} & \noheatcell[0]{181} & \noheatcell[1]{186} & \noheatcell[2]{202} & \noheatcell[1]{227} & \noheatcell[3]{455} & \noheatcell[6]{350} & \noheatcell[0]{246} & \noheatcell[6]{396} & \noheatcell[7]{489} & \noheatcell[5]{493} & \noheatcell[0]{227} & \noheatcell[2]{321} & \textbf{\noheatcell{284}} \\
& Cadrille
& \noheatcell[1]{76} & \noheatcell[1]{86} & \noheatcell[2]{114} & \noheatcell[2]{175} & \noheatcell[3]{141} & \noheatcell[4]{174} & \noheatcell[7]{319} & \noheatcell[4]{329} & \noheatcell[2]{190} & \noheatcell[6]{316} & \noheatcell[5]{262} & \noheatcell[1]{358} & \noheatcell[0]{158} & \noheatcell[2]{318} & \textbf{\noheatcell{220}} \\
\cmidrule(lr){2-17}
& Claude Opus
& \noheatcell[1]{57} & \noheatcell[1]{55} & \noheatcell[4]{84} & \noheatcell[4]{101} & \noheatcell[1]{104} & \noheatcell[3]{128} & \noheatcell[3]{317} & \noheatcell[5]{282} & \noheatcell[3]{161} & \noheatcell[3]{280} & \noheatcell[3]{245} & \noheatcell[3]{323} & \noheatcell[2]{133} & \noheatcell[24]{269} & \textbf{\noheatcell{186}} \\
& Gemini
& \noheatcell[0]{58} & \noheatcell[0]{58} & \noheatcell[1]{80} & \noheatcell[2]{97} & \noheatcell[1]{99} & \noheatcell[2]{124} & \noheatcell[4]{288} & \noheatcell[9]{258} & \noheatcell[1]{147} & \noheatcell[7]{280} & \noheatcell[7]{251} & \noheatcell[5]{318} & \noheatcell[2]{147} & \noheatcell[4]{260} & \textbf{\noheatcell{182}} \\
& GPT 5.4
& \noheatcell[0]{76} & \noheatcell[1]{79} & \noheatcell[2]{103} & \noheatcell[3]{123} & \noheatcell[2]{128} & \noheatcell[2]{153} & \noheatcell[18]{370} & \noheatcell[7]{306} & \noheatcell[1]{189} & \noheatcell[11]{353} & \noheatcell[4]{266} & \noheatcell[8]{354} & \noheatcell[2]{163} & \noheatcell[5]{299} & \textbf{\noheatcell{216}} \\
& Kimi K2.6
& \noheatcell[1]{63} & \noheatcell[0]{60} & \noheatcell[1]{86} & \noheatcell[2]{105} & \noheatcell[0]{114} & \noheatcell[5]{136} & \noheatcell[18]{397} & \noheatcell[6]{305} & \noheatcell[4]{182} & \noheatcell[12]{357} & \noheatcell[3]{305} & \noheatcell[17]{426} & \noheatcell[1]{121} & \noheatcell[5]{347} & \textbf{\noheatcell{219}} \\
& Qwen 27B
& \noheatcell[0]{73} & \noheatcell[2]{69} & \noheatcell[1]{90} & \noheatcell[2]{95} & \noheatcell[2]{116} & \noheatcell[5]{126} & \noheatcell[10]{234} & \noheatcell[8]{210} & \noheatcell[1]{151} & \noheatcell[7]{239} & \noheatcell[3]{202} & \noheatcell[16]{255} & \noheatcell[3]{119} & \noheatcell[2]{170} & \textbf{\noheatcell{151}} \\
& Qwen 9B
& \noheatcell[4]{72} & \noheatcell[1]{65} & \noheatcell[5]{94} & \noheatcell[6]{89} & \noheatcell[3]{128} & \noheatcell[6]{135} & \noheatcell[12]{189} & \noheatcell[24]{220} & \noheatcell[5]{146} & \noheatcell[15]{164} & \noheatcell[24]{194} & \noheatcell[32]{238} & \noheatcell[8]{99} & \noheatcell[6]{172} & \textbf{\noheatcell{141}} \\
& CADCoder
& \noheatcell{131} & \noheatcell{125} & \noheatcell{174} & \noheatcell{185} & \noheatcell{202} & \noheatcell{204} & \noheatcell{586} & \noheatcell{444} & \noheatcell{312} & \noheatcell{521} & \noheatcell{598} & \noheatcell{600} & \noheatcell{244} & \noheatcell{580} & \textbf{\noheatcell{364}} \\
\midrule
\multirow{11}{*}{\textbf{Operation Count}}
& CADFit
& \noheatcell{12} & \noheatcell{13} & \noheatcell{15} & \noheatcell{19} & \noheatcell{18} & \noheatcell{23} & \noheatcell{66} & \noheatcell{42} & \noheatcell{24} & \noheatcell{59} & \noheatcell{82} & \noheatcell{73} & \noheatcell{32} & \noheatcell{72} & \textbf{\noheatcell{42}} \\
& CAD-Recode
& \noheatcell[1]{9} & \noheatcell[0]{9} & \noheatcell[1]{12} & \noheatcell[0]{19} & \noheatcell[0]{15} & \noheatcell[1]{18} & \noheatcell[1]{38} & \noheatcell[1]{36} & \noheatcell[0]{21} & \noheatcell[1]{36} & \noheatcell[1]{30} & \noheatcell[1]{41} & \noheatcell[0]{19} & \noheatcell[1]{36} & \textbf{\noheatcell{25}} \\
& CADEvolve
& \noheatcell[0]{9} & \noheatcell[0]{9} & \noheatcell[0]{15} & \noheatcell[0]{15} & \noheatcell[1]{17} & \noheatcell[0]{18} & \noheatcell[0]{34} & \noheatcell[0]{28} & \noheatcell[0]{20} & \noheatcell[1]{32} & \noheatcell[0]{26} & \noheatcell[0]{36} & \noheatcell[0]{18} & \noheatcell[0]{25} & \textbf{\noheatcell{22}} \\
& Cadrille
& \noheatcell[0]{8} & \noheatcell[0]{9} & \noheatcell[0]{12} & \noheatcell[0]{18} & \noheatcell[1]{15} & \noheatcell[0]{18} & \noheatcell[1]{34} & \noheatcell[1]{35} & \noheatcell[1]{20} & \noheatcell[1]{33} & \noheatcell[1]{27} & \noheatcell[1]{38} & \noheatcell[0]{17} & \noheatcell[0]{34} & \textbf{\noheatcell{23}} \\
\cmidrule(lr){2-17}
& Claude Opus
& \noheatcell[0]{6} & \noheatcell[0]{6} & \noheatcell[0]{7} & \noheatcell[1]{10} & \noheatcell[1]{8} & \noheatcell[0]{11} & \noheatcell[1]{20} & \noheatcell[0]{22} & \noheatcell[0]{13} & \noheatcell[0]{21} & \noheatcell[1]{12} & \noheatcell[1]{22} & \noheatcell[0]{13} & \noheatcell[2]{21} & \textbf{\noheatcell{14}} \\
& Gemini
& \noheatcell[0]{5} & \noheatcell[0]{5} & \noheatcell[0]{6} & \noheatcell[0]{9} & \noheatcell[0]{7} & \noheatcell[1]{10} & \noheatcell[0]{16} & \noheatcell[1]{18} & \noheatcell[1]{11} & \noheatcell[1]{19} & \noheatcell[1]{12} & \noheatcell[1]{21} & \noheatcell[1]{11} & \noheatcell[1]{18} & \textbf{\noheatcell{13}} \\
& GPT 5.4
& \noheatcell[0]{6} & \noheatcell[0]{6} & \noheatcell[0]{7} & \noheatcell[1]{10} & \noheatcell[0]{9} & \noheatcell[0]{11} & \noheatcell[0]{21} & \noheatcell[0]{22} & \noheatcell[0]{14} & \noheatcell[1]{23} & \noheatcell[1]{13} & \noheatcell[1]{23} & \noheatcell[0]{13} & \noheatcell[1]{22} & \textbf{\noheatcell{15}} \\
& Kimi K2.6
& \noheatcell[0]{6} & \noheatcell[0]{6} & \noheatcell[1]{8} & \noheatcell[1]{11} & \noheatcell[0]{11} & \noheatcell[1]{12} & \noheatcell[1]{28} & \noheatcell[1]{25} & \noheatcell[0]{16} & \noheatcell[1]{28} & \noheatcell[1]{18} & \noheatcell[0]{28} & \noheatcell[0]{13} & \noheatcell[1]{28} & \textbf{\noheatcell{18}} \\
& Qwen 27B
& \noheatcell[0]{7} & \noheatcell[0]{7} & \noheatcell[1]{8} & \noheatcell[1]{9} & \noheatcell[0]{10} & \noheatcell[0]{11} & \noheatcell[1]{17} & \noheatcell[1]{18} & \noheatcell[1]{13} & \noheatcell[1]{18} & \noheatcell[0]{12} & \noheatcell[1]{17} & \noheatcell[0]{11} & \noheatcell[1]{15} & \textbf{\noheatcell{12}} \\
& Qwen 9B
& \noheatcell[0]{7} & \noheatcell[0]{7} & \noheatcell[0]{9} & \noheatcell[0]{9} & \noheatcell[2]{14} & \noheatcell[1]{14} & \noheatcell[1]{15} & \noheatcell[3]{21} & \noheatcell[0]{15} & \noheatcell[1]{15} & \noheatcell[3]{14} & \noheatcell[3]{18} & \noheatcell[1]{10} & \noheatcell[1]{16} & \textbf{\noheatcell{13}} \\
& CADCoder
& \noheatcell{15} & \noheatcell{13} & \noheatcell{21} & \noheatcell{22} & \noheatcell{24} & \noheatcell{24} & \noheatcell{68} & \noheatcell{52} & \noheatcell{38} & \noheatcell{62} & \noheatcell{69} & \noheatcell{71} & \noheatcell{29} & \noheatcell{67} & \textbf{\noheatcell{43}} \\
\bottomrule
\end{tabular}%
}
\end{table}

\subsubsection{Inference Speed Comparisons}
\label{app:inference_speed}
In this section, we compare the inference speeds of the evaluated models. Table \ref{tab:inference_speed} details the time required to generate CADQuery code for each sample across the different methods. To ensure a fair comparison, all local models are run on a standardized setup featuring an NVIDIA RTX PRO 6000 GPU and an AMD Ryzen Threadripper PRO 7975WX CPU. We evaluate the models across the entire benchmark suite using the authors' original code, without applying any supplementary optimizations. For the large frontier models, we report API response times using single-view image inputs on a random subset of 100 samples. As the results show, CADFit requires significantly more inference time compared to the deep learning-based methods.

\begin{table}[H]
\centering
\small
\setlength{\tabcolsep}{3.2pt}
\renewcommand{\arraystretch}{1.08}
\caption{\textbf{Inference time metrics} (in seconds) across evaluated methods. Local models were tested on the complete benchmark suite using standardized hardware. For frontier models, metrics reflect API response times over a random subset of 100 samples.}
\label{tab:inference_speed}
\resizebox{\textwidth}{!}{%
\begin{tabular}{l cccc | ccccccc}
\toprule
Metric & CADFit & CAD-Recode & CADEvolve & Cadrille & Claude Opus & Gemini & GPT 5.4 & Kimi K2.6 & Qwen 27B & Qwen 9B & CADCoder \\
\midrule
Mean    & 453.1 & 2.095 & 1.491 & 3.262 & 6.870 & 30.388 & 7.478 & 41.089 & 44.566 & 19.575 & 81.058 \\
Median  & 291.8 & 1.751 & 0.0631 & 3.070 & 5.743 & 21.264 & 6.006 & 17.620 & 42.429 & 19.067 & 64.847 \\
Std     & 540.2 & 1.460 & 2.715 & 1.286 & 3.779 & 45.554 & 5.127 & 80.087 & 17.910 & 4.540 & 57.201\\
Minimum & 37.3 & 0.464 & 0.0516 & 1.212 & 1.787 & 4.033 & 1.699 & 4.676 & 21.058 & 12.251 & 12.750 \\
Maximum & 6817.6 & 7.754 & 13.22 & 8.937 & 18.177 & 410.823 & 23.811 & 612.8 & 150.342 & 46.037 & 205.186 \\
\bottomrule
\end{tabular}%
}
\end{table}

\subsection{Understanding the Effects of Complexity and Diversity Sampling}

In this section, we further investigate the impact of including complexity-stratification and diversity-aware sampling in our benchmark splits.

\subsubsection{Does face count measure difficulty when operation vocabulary is expanded?}
\label{app:complexity_strat}

\begin{figure}[H]
\includegraphics[width=1\linewidth]{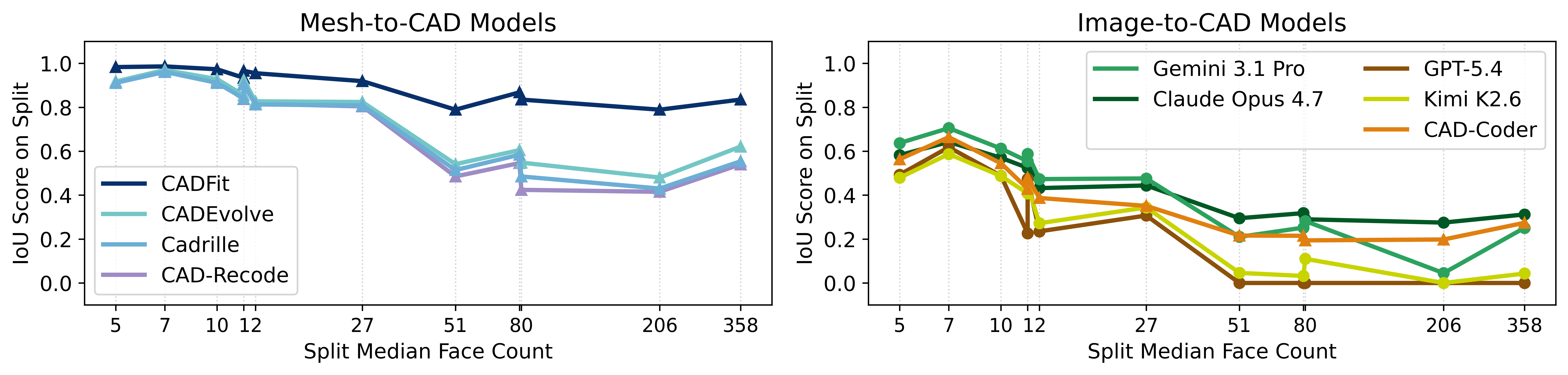}
\caption{Model performance as a function of face count when all STEP-derived benchmark families (B, F, E, A) are included. The y-axis shows median IoU score for a split, while the x-axis shows split median face count on a logarithmic scale. In contrast to the controlled sketch-and-extrude analysis in Figure~\ref{fig:complexity_analysis}, the relationship between face count and reconstruction performance is less monotonic when Fusion and All-Ops splits are included. This suggests that face count is a useful proxy for complexity within controlled operation families (e.g., sketch-and-extrude only), but does not fully capture reconstruction difficulty when operation vocabulary varies more broadly.}
\label{fig:complexity_analysis_all_ops}
\end{figure}

\subsubsection{Does DINOv3 diversity sampling matter? — Diversity-selected versus random evaluation}
\label{app:diversity_sampling}

To assess the effect of diversity-aware sampling on measured model performance, we construct randomly sampled variants of the Base-Low, Base-Medium, Extrude-Low, and Extrude-Medium splits while preserving the same face-count ranges used for complexity stratification. We then evaluate \cadevolve on both the diversity-sampled \benchmark splits and the corresponding random splits (Table~\ref{tab:cadevolve_diversity_comparison}). Across all four splits, diversity-selected samples yield lower median IoU than randomly sampled examples, suggesting that diversity-aware sampling can expose more challenging evaluation cases within the same nominal complexity range. The effect is modest for Base-Low, Base-Medium, and Extrude-Low, but is larger for Extrude-Medium, where IoU decreases by 11.33\%. This larger drop may indicate that random sampling in this range includes more visually or geometrically redundant examples, while DINOv3-based sampling selects a more varied set of sketch-and-extrude geometries.  

\begin{table}[H]
\centering
\small
\caption{Comparison of \cadevolve median IoU on Base and Extrude splits with and without diversity-aware sampling. Percent change is computed relative to the randomly sampled split.}
\label{tab:cadevolve_diversity_comparison}
\begin{tabular}{lccc}
\toprule
Split & \makecell{Diversity Sampled \\ (\benchmark)} & Randomly Sampled & \% Change \\
\midrule
Base-Low & 0.970 & 0.984 & -1.42\% \\
Base-Medium & 0.924 & 0.934 & -1.07\% \\
Extrude-Low & 0.928 & 0.929 & -0.11\% \\
Extrude-Medium & 0.548 & 0.618 & -11.33\% \\
\bottomrule
\end{tabular}
\end{table}

\subsection{Performance Deltas Across Input Modalities}
\label{app:modality_shift}
The following tables report performance deltas across input modalities, providing a more detailed view of model robustness to mesh noise, photorealistic rendering, multi-view inputs, and standardized single-view render formats.

\begin{table}[H]
\centering
\small
\caption{\textbf{Robustness of mesh-to-CAD models to noisy mesh inputs.}
We report aggregate IoU and VSR across CADBench splits for clean and noisy mesh inputs. 
$\Delta$ denotes the change from clean to noisy inputs, with negative values indicating degradation.}
\label{tab:mesh_modality_delta}
\begin{tabular}{lcccccc}
\toprule
Method 
& Clean IoU & Noisy IoU & $\Delta$ IoU 
& Clean VSR & Noisy VSR & $\Delta$ VSR \\
\midrule
CADFit & 0.859 & 0.808 & \deltacell{-0.051} & 0.981 & 0.998 & \deltacell{+0.017} \\
CADEvolve & 0.707 & 0.130 & \deltacell{-0.577} & 0.967 & 0.991 & \deltacell{+0.024} \\
Cadrille & 0.687 & 0.579 & \deltacell{-0.108} & 0.939 & 0.914 & \deltacell{-0.025} \\
CAD-Recode & 0.673 & 0.511 & \deltacell{-0.162} & 0.912 & 0.868 & \deltacell{-0.044} \\
\bottomrule
\end{tabular}
\end{table}

\begin{table}[H]
\centering
\small
\caption{\textbf{Robustness of image-to-CAD models across varying image modalities.}
We report aggregate IoU across \benchmark splits and modalities: single-view (SV), photorealistic (Photo), and multi-view (MV) renders. Deltas are computed relative to IoU on single-view inputs.}
\label{tab:image_modality_delta}
\begin{tabular}{lccccc}
\toprule
Method 
& SV IoU & Photo IoU & $\Delta_{\mathrm{photo}}$ 
& MV IoU & $\Delta_{\mathrm{multi}}$ \\
\midrule
Gemini 3.1 Pro & 0.382 & 0.388 & \deltacell{+0.006} & 0.374 & \deltacell{-0.008} \\
Claude Opus 4.7 & 0.412 & 0.409 & \deltacell{-0.003} & 0.418 & \deltacell{+0.006} \\
GPT-5.4 & 0.158 & 0.145 & \deltacell{-0.013} & 0.153 & \deltacell{-0.005} \\
Kimi K2.6 & 0.224 & 0.196 & \deltacell{-0.028} & 0.131 & \deltacell{-0.093} \\
CAD-Coder & 0.354 & 0.265 & \deltacell{-0.089} & 0.159 & \deltacell{-0.195} \\
Qwen3.5 27B & 0.015 & 0.017 & \deltacell{+0.002} & 0.000 & \deltacell{-0.015} \\
Qwen3.5 9B & 0.000 & 0.000 & \deltacell{+0.000} & 0.000 & \deltacell{+0.000} \\
\bottomrule
\end{tabular}
\end{table}

\begin{table}[H]
\centering
\small
\caption{\textbf{Executability robustness across image modalities.}
We report aggregate valid shape rate (VSR) across CADBench for different modalities: single-view (SV), photorealistic (Photo), and multi-view (MV). Deltas are computed relative to VSR on single-view  inputs.}
\label{tab:image_modality_vsr_delta}
\begin{tabular}{lccccc}
\toprule
Method 
& SV VSR & Photo VSR & $\Delta_{\mathrm{photo}}$ 
& MV VSR & $\Delta_{\mathrm{multi}}$ \\
\midrule
Gemini 3.1 Pro & 0.729 & 0.759 & \deltacell{+0.030} & 0.740 & \deltacell{+0.011} \\
Claude Opus 4.7 & 0.798 & 0.821 & \deltacell{+0.023} & 0.797 & \deltacell{-0.001} \\
GPT-5.4 & 0.513 & 0.529 & \deltacell{+0.016} & 0.499 & \deltacell{-0.014} \\
Kimi K2.6 & 0.616 & 0.628 & \deltacell{+0.012} & 0.548 & \deltacell{-0.068} \\
CAD-Coder & 0.946 & 0.948 & \deltacell{+0.002} & 0.903 & \deltacell{-0.043} \\
Qwen3.5 27B & 0.335 & 0.353 & \deltacell{+0.018} & 0.271 & \deltacell{-0.064} \\
Qwen3.5 9B & 0.099 & 0.091 & \deltacell{-0.008} & 0.034 & \deltacell{-0.065} \\
\bottomrule
\end{tabular}
\end{table}

\begin{table}[H]
\centering
\small
\caption{\textbf{Effect of standardized image inputs on models with mesh- and image-conditioned inference modes.}
We compare aggregate IoU and VSR when models are evaluated using their mesh-conditioned pipeline versus \benchmark's standardized single-view (SV) renders. $\Delta$ denotes the change from mesh-conditioned to SV-conditioned inputs, with negative values indicating degradation. For both Cadrille and CADEvolve, switching from mesh-conditioned inputs to our standardized single-view renders causes IoU to drop effectively to zero.}
\label{tab:mesh_image_conditioned}
\begin{tabular}{lcccccc}
\toprule
Method 
& Mesh IoU & SV IoU & $\Delta$ IoU 
& Mesh VSR & SV VSR & $\Delta$ VSR \\
\midrule
Cadrille  & 0.687 & 0.039 & \deltacell{-0.648} & 0.939 & 0.788 & \deltacell{-0.151} \\
CADEvolve & 0.707 & 0.097 & \deltacell{-0.610} & 0.967 & 0.960 & \deltacell{-0.007} \\
\bottomrule
\end{tabular}
\end{table}

\subsection{Are the Metrics Diagnostic?}

\subsubsection{Are IoU, Chamfer distance, and SIoU redundant? — Metric correlation analysis}
\label{app:metrics_redundant}
To assess whether our geometric fidelity metrics provide complementary information, we compute sample-level Spearman correlations across approximately 1M syntactically valid model-generated CAD programs and visualize the corresponding pairwise metric relationships. IoU and SIoU are only moderately correlated, while CD is negatively correlated with both higher-is-better metrics, indicating that these metrics are related but not redundant.

\begin{figure}[!ht]
\centering
\includegraphics[width=0.45\linewidth]{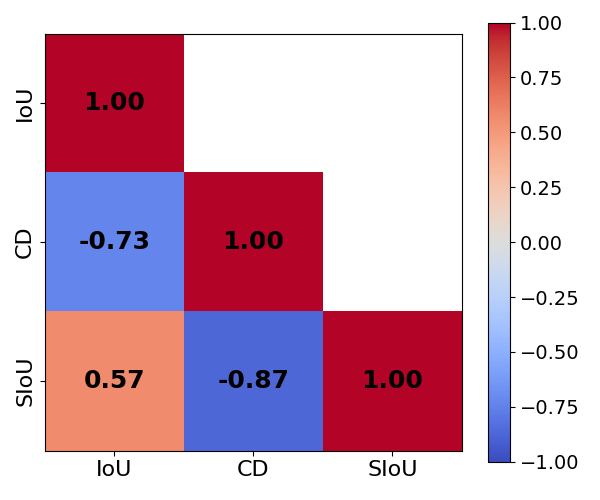}
\caption{\textbf{Sample-level correlations between geometric fidelity metrics.} Spearman rank correlations are computed across approximately 1M syntactically valid model-generated CAD programs. IoU and SIoU are moderately correlated, while CD is negatively correlated with both higher-is-better metrics.}
\label{fig:correlations}
\end{figure}

\begin{figure}[!ht]
\centering
\includegraphics[width=1.0\linewidth]{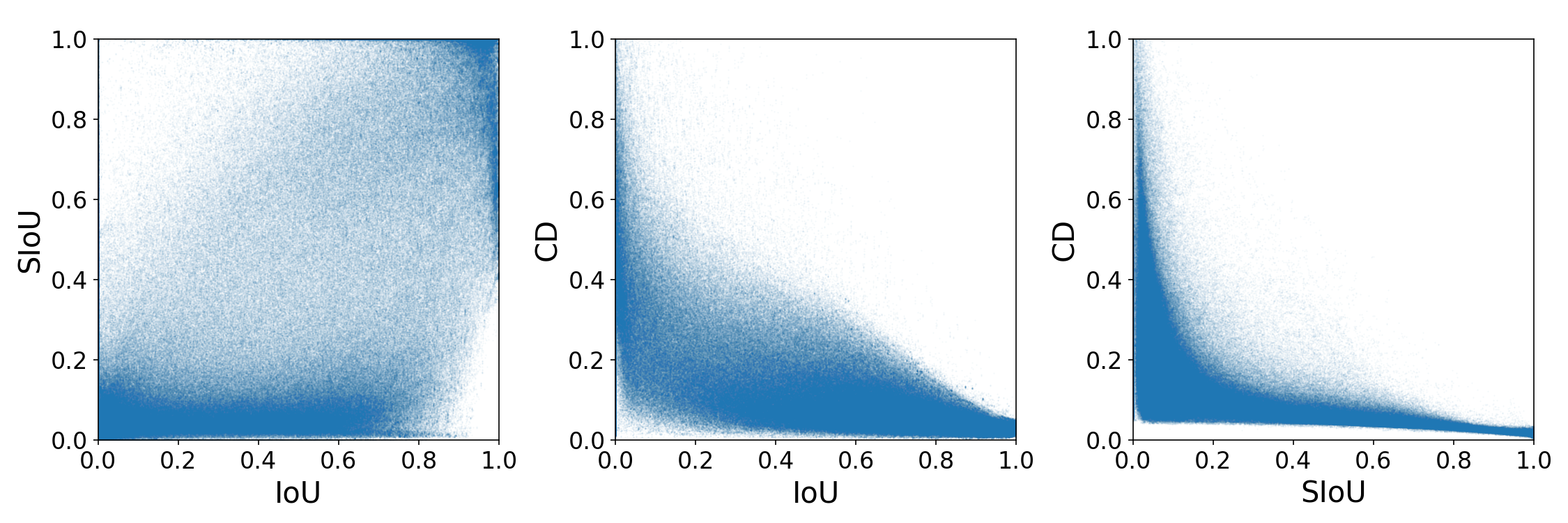}
\caption{\textbf{Pairwise relationships between geometric fidelity metrics.} Each point corresponds to one syntactically valid model-generated CAD program from the full set of approximately 1M predictions. In particular, the broad scatter between IoU and SIoU shows that volumetric overlap and thresholded surface coverage can differ substantially at the sample level.}
\label{fig:metric_pairwise_scatter}
\end{figure}

\subsubsection{When is one geometric fidelity metric not enough? — High-IoU/low-SIoU and high-SIoU/low-IoU examples}
\label{app:visual_metrics}
\begin{figure}[!ht]
\centering
\includegraphics[width=1.0\linewidth]{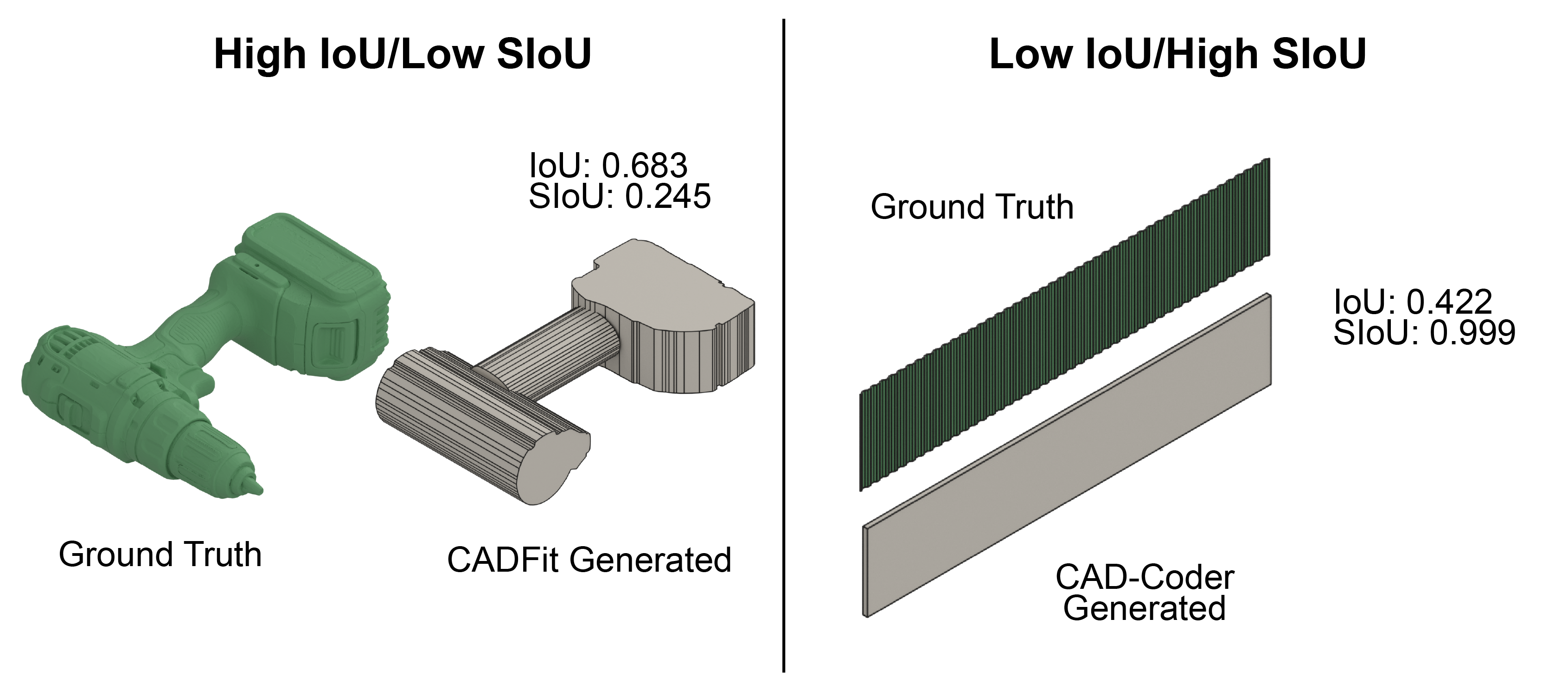}
\caption{\textbf{Examples where IoU and SIoU disagree.} High IoU without computing SIoU can obscure missing surface detail, while high SIoU with low IoU can occur when predicted surfaces align with the target but the enclosed volume is incorrect.}
\label{fig:iou_siou}
\end{figure}

Figure~\ref{fig:iou_siou} visualizes two cases where IoU and SIoU disagree. In the high-IoU/low-SIoU case, the generated solid recovers the target's overall volume but misses fine surface structure, leading to poor surface coverage despite strong volumetric overlap. In the low-IoU/high-SIoU case, the generated model places surfaces near the target geometry but fails to recover the correct enclosed volume, for example due to incorrect thickness. These examples illustrate why volumetric and surface-based metrics are complementary rather than interchangeable.

\section{Asset Release, Dataset Licensing, and Broader Impacts}
\label{app:license}

The \benchmark comprises data from the following sources, each governed by its own licensing terms:

\begin{itemize}[leftmargin = *]
    \item \textbf{DeepCAD:} Licensed under the MIT License. 
    \item \textbf{Fusion 360 Gallery:} The dataset and supporting codebase are publicly available on GitHub\footnote{\url{https://github.com/AutodeskAILab/Fusion360GalleryDataset}} under a license permitting non-commercial research, mirroring the terms of the ImageNet \cite{deng2009imagenet} license.
    
    \item \textbf{ABC:} Licensed under the MIT License.
    \item \textbf{Objaverse:} The dataset is collectively licensed under ODC-By v1.0, while individual objects are distributed as Creative Commons assets under various specific licenses.\footnote{Detailed license information is available at \url{https://huggingface.co/datasets/allenai/objaverse}.}
    \item \textbf{MCB:} Licensed under the MIT License.
\end{itemize}

\textbf{Access-controlled datasets:} CC3D is an important and widely used scan-to-CAD benchmark, and we include it in our discussion of related datasets and prior evaluation settings. However, we do not include CC3D in the primary CADBench evaluation suite or released aggregate benchmark because access to the dataset is governed by a separate institutional data-use agreement that must be requested and executed by prospective users. In contrast, CADBench is designed around artifacts that can be directly inspected, reproduced, and rerun by reviewers and future users without requiring additional third-party licensing negotiations. For this reason, CADBench does not report quantitative results on CC3D. This exclusion reflects a reproducibility and release-design consideration rather than any judgment about the scientific value or importance of CC3D as a benchmark.

\paragraph{Broader impacts and responsible use:} CADBench is intended to support progress toward more reliable AI-assisted CAD tools by providing standardized evaluation of geometric fidelity, executability, and program compactness across different modalities and benchmark families. CAD-generating AI tools could reduce the time and expertise required to create editable engineering models and improve access to CAD workflows. However, AI CAD generation systems could also be misused to generate unsafe or poorly validated designs, or could lead users to over-trust generated CAD programs in safety-critical engineering contexts. CADBench is an evaluation benchmark rather than a deployed design system, and benchmark performance should not be interpreted as certification that generated CAD models are suitable for manufacturing or real-world use. We recommend that any generated CAD models be reviewed by qualified engineers and validated with domain-appropriate simulation, analysis, and safety checks before fabrication or deployment.

\newpage

\end{document}